\documentclass[pdflatex,sn-mathphys-num,iicol]{sn-jnl}

\usepackage{graphicx}%
\usepackage{multirow}%
\usepackage{amsmath,amssymb,amsfonts}%
\usepackage{amsthm}%
\usepackage{mathrsfs}%
\usepackage[title]{appendix}%
\usepackage{xcolor}%
\usepackage{textcomp}%
\usepackage{manyfoot}%
\usepackage{booktabs}%
\usepackage{algorithm}%
\usepackage{algorithmicx}%
\usepackage{algpseudocode}%
\usepackage{listings}%
\usepackage{amsthm,amsmath,amssymb,lipsum}
\usepackage{mathrsfs}
\usepackage{natbib}

\usepackage{multirow}
\usepackage{booktabs}
\usepackage{tabularx}
\usepackage{makecell}
\usepackage{adjustbox}
\usepackage{array}
\usepackage{tabulary}
\usepackage{colortbl}
\usepackage{pifont} 
\usepackage{adjustbox}
\usepackage{color, soul}
\sethlcolor{yellow}
\soulregister\cite7
\soulregister\citet7
\soulregister\citep7
\soulregister\ref7
\soulregister\underline7
\soulregister\textcolor7

\AtBeginEnvironment{algorithm}{\small}
\algrenewcommand\alglinenumber[1]{\scriptsize #1}
\algrenewcommand\algorithmicindent{1.2em}

\algtext*{EndFor}
\algtext*{EndIf}

\newcolumntype{I}{!{\vrule width 3pt}}
\newlength\savedwidth

\newlength\savewidth
\newcommand\shline{\noalign{\global\savewidth\arrayrulewidth
\global\arrayrulewidth 1.25pt}%
\hline
\noalign{\global\arrayrulewidth\savewidth}}

\theoremstyle{thmstyleone}%

\theoremstyle{thmstyletwo}%

\theoremstyle{thmstylethree}%

\begin{document}

\title[Article Title]{Let Synthetic Data Shine: Domain Reassembly and Soft-Fusion for Single Domain Generalization}

\author[1]{\fnm{Hao} \sur{Li}}
\email{fantasioly@gmail.com}

\author[2]{\fnm{Yubin} \sur{Xiao}}
\equalcont{These authors contributed equally to this work.}

\author[1]{\fnm{Ke } \sur{Liang}}

\author[1]{\fnm{Mengzhu} \sur{Wang}}

\author*[1]{\fnm{Long } \sur{Lan}}
\email{long.lan@nudt.edu.cn}

\author[3]{\fnm{Kenli } \sur{Li}}
\email{lkl@hnu.edu.cn}

\author[1]{\fnm{Xinwang } \sur{Liu}}
\email{xinwangliu@nudt.edu.cn}

\affil[1]{\orgdiv{College of Computer Science and Technology}, \orgname{National University of Defense Technology}, \city{Changsha}, \postcode{410073},  \country{China}}

\affil[2]{\orgdiv{College of Computer Science and Technology}, \orgname{ Jilin University}, \orgaddress{\city{Changchun}, \postcode{130012}, \country{China}}}

\affil[3]{\orgdiv{College of Computer Science and Electronic Engineering}, \orgname{Hunan University}, \orgaddress{\city{Changsha}, \postcode{410082}, \country{China}}}

\abstract{
Single Domain Generalization (SDG) aims to train models that maintain consistent performance across diverse scenarios using data from a single source. While latent diffusion models (LDMs) show promise for augmenting limited source data, our analysis reveals that directly employing synthetic data may not only fail to provide benefits but can actually compromise performance due to substantial feature distribution discrepancies between synthetic and real target domains.
To address this issue, we propose Discriminative Domain Reassembly and Soft-Fusion (DRSF), a training framework leveraging synthetic data to improve model generalization. We employ LDMs to produce diverse pseudo-target domain samples and introduce two key modules to handle distribution bias. First, Discriminative Feature Decoupling and Reassembly (DFDR) module uses entropy-guided attention to recalibrate channel-level features, suppressing synthetic noise while preserving semantic consistency. Second, Multi-pseudo-domain Soft Fusion (MDSF) module uses adversarial training with latent-space feature interpolation, creating continuous feature transitions between domains.
Extensive SDG experiments on image classification, object detection, and semantic segmentation demonstrate that DRSF delivers substantial performance gains with only marginal computational overhead. Notably, DRSF's plug-and-play architecture enables seamless integration with unsupervised domain adaptation paradigms, underscoring its broad applicability to diverse, real-world domain challenges.
}
\maketitle

\begin{figure*}[!t]
\begin{center}
    \centering
    \includegraphics[width=\textwidth]{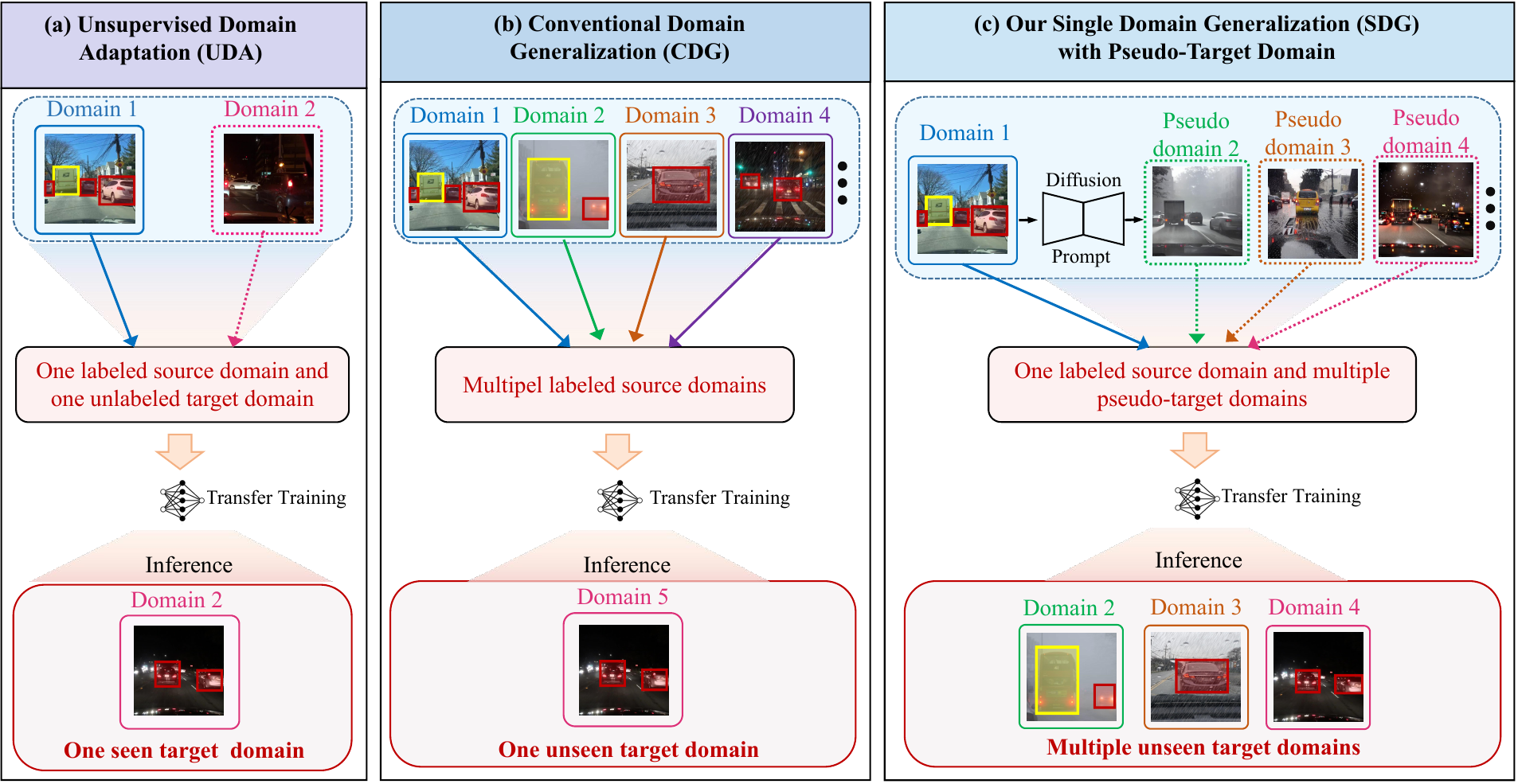}
    \caption{A comparison of Domain Adaptation (DA), Conventional Domain Generalization (CDG), and our pseudo-target-domain-based Single Domain Generalization (SDG). DA relies on aligning with target-domain data (left), whereas CDG requires joint training across multiple source domains (middle). In contrast, our approach requires only single-source data, using diffusion models to generate diverse pseudo-target domains (right).
}
    \label{fig:sdg_task}
\end{center}
\end{figure*}

\section{Introduction}\label{sec1}

Supervised and semi-supervised deep learning methods have made significant advances in the field of computer vision (CV) \cite{lecun2015deep,ren2015fasterrcnn}. However, these approaches are often based on the assumption that the training (source) and testing (target) data share the same distribution. In the presence of domain shift \cite{wang2025deep}, the model's performance in target domains can degrade sharply, severely limiting its practical applicability in real-world scenarios \cite{zhou2021domain}. To address the domain shift problem, numerous methods in Unsupervised Domain Adaptation (UDA) \cite{ben2010theory,chen2021sada,jiang2021dadapt} and Domain Generalization (DG) \cite{bellitto2021hierarchical,du2022cross,zhou2023semi,SHADE} have been proposed. UDA focuses on transferring models trained on source domain to (accessible) target domains, whereas DG aims to develop models that exhibit strong cross-domain generalization capabilities across multiple (unseen) target domains. Comparing to UDA, DG emphasizes the development of plug-and-play models that can generalize effectively without the need for additional training or fine-tuning, garnering increasing attention \cite{zhou2022domain}. More importantly, in real-world applications, researchers often encounter constraints related to single-source domain data (Single Domain Generalization, SDG) \cite{yuan2023domain}, where only data from a single source domain are accessible. This paper specifically focuses on this challenging yet realistic problem. Additionally, we illustrate the differences between UDA, DG, and our proposed synthetic data-based method to solve SDG challenges in Fig.~\ref{fig:sdg_task}.

Existing methods in SDG primarily rely on data augmentation techniques to diversify source domain data, which has led to some progress in image classification tasks \cite{su2023rethinking,sdgda}. However, data augmentation may introduce noise; improper image augmentation may result in the loss of important image edge information, causing distortions in the target structure and deviating from the original (pre-augmentation) samples \cite{termohlen2023re,liu2024unbiased}. This complicates the application of these methods to broader CV tasks, such as object detection. More importantly, constrained by the limited representation of a single source domain, even augmented data tends to cause models to overfit to the specific features of the source domain. This limits the representation of training data, makes it difficult to simulate the diversity of target domains in real-world scenarios, and ultimately reduces the generalization ability of the model \cite{liu2024unbiased}.

Recently, generative diffusion models have become a crucial tool for enhancing training data representations due to their high fidelity and versatile cross-domain generation capabilities \cite{schuhmann2022laion}. By leveraging these generative models to synthesize samples across different scenarios, it is possible to both mitigate the lack of diversified source domain data and expose the model to a wide range of potential target domains during training, thereby improving its generalization ability in unseen domains. Consequently, using generative models to enhance  model generalization has emerged as a promising approach in SDG \cite{vidit2023clip}. However, existing methods primarily focus on improving the visual realism of synthetic data, often overlooking semantic consistency and distribution transferability between the synthetic domain and the real target domain. Direct application of synthetic data may fail to enhance model generalization effectively and could even lead to degeneration, which limits the practical utility of the synthetic data (see Section~\ref{sec:challenge} for details). Therefore, determining how to effectively utilize the synthetic data, which inherently carries distribution biases, to improve the model’s generalization ability becomes a key challenge in the field of SDG.

To address the aforementioned challenge, we propose a novel and general training framework, DRSF (Discriminative Domain Reassembly and Soft-Fusion), which utilizes synthetic data to improve generalization in SDG. We employ a generative diffusion model to produce rich and diverse target domain samples, thereby enhancing the diversity of the training data. Notably, while we also utilize synthetic data, our approach differs from existing methods by introducing two core modules designed to handle the distribution bias in synthetic data. Firstly, to mitigate the significant feature distribution gap between the synthetic pseudo-target domain data and the source domain data, which could cause model confusion during training, we propose a Discriminative Feature Decoupling and Reassembly (DFDR) module to decompose the features of both domains into primary features (domain-invariant) and shared features (domain-specific). We then reduce content discrepancy by aligning the primary features, while using the shared features to capture style variations across domains, thus enhancing the model's cross-domain generalization ability. Secondly, to further improve the model's adaptability to unseen target domains, we propose a Multi-Pseudo-domain Feature Soft-Fusion (MDSF) module. By blending the features of the source domain and multiple pseudo-target domains in the shared feature space, MDSF constructs a continuous domain-invariant feature space to smooth the boundaries between different domains and combines adversarial training to further narrow the distribution difference between the source domain and pseudo-target domain, thus fusing the decoupled features. It is important to note that, beyond SDG, we also demonstrate the extension of DRSF to UDA tasks, which further highlights the plug-and-play generality of our approach and its practical applicability to a wide range of tasks in real-world scenarios.

The key contributions of this work are as follows.

{\romannumeral 1}) We demonstrate that although the synthetic data visually resembles real data, their internal distribution differs significantly, hampering the effective enhancement of the model’s SDG generalization when using the synthetic data directly and potentially leading to localized degradation. To the best of our knowledge, our work is the first one reporting this finding in the SDG field with the support of experimental results.

{\romannumeral 2}) To effectively leverage synthetic data with inherent distributional biases, we introduce DRSF--a plug-and-play training framework designed to enhance SDG performance by decoupling and recombining features from synthetic data. To the best of our knowledge, DRSF is the first general SDG training framework that manipulates synthetic data at the feature level.

{\romannumeral 3}) We demonstrate the state-of-the-art (SOTA) generalization capabilities of DRSF in SDG tasks by conducting extensive experiments on object detection and semantic segmentation. Additionally, ablation studies validate the effectiveness of the proposed DFDR and MDSF modules.

\section{Related work}
In this section, we first review UDA and DG methods, followed by an introduction to SDG methods applied in object detection and semantic segmentation. Finally, we present pioneering works that employs synthetic data to address SDG challenges. 

\subsection{Unsupervised Domain Adaptation and Domain Generalization}

In CV tasks, models typically assume that the source and target data share the same distribution. However, significant real-world distribution differences often cause performance degradation in the target domain—a phenomenon known as domain shift \cite{vapnik1991principles}. For instance, in tasks such as image classification, object detection, and semantic segmentation, these disparities can stem from variations in environments, illumination conditions, or viewing angles \cite{wang2025deep,yang2024learning}.

To address this challenge, UDA and DG have emerged as two pivotal techniques. As illustrated in Fig.~\ref{fig:sdg_task}(a), UDA transfers knowledge by adapting to unlabeled target data, typically employing methods such as adversarial training \cite{zhao2021madan,munir2023domain,qu2024aamt} or maximum mean discrepancy minimization \cite{wang2021rethinking}. However, UDA's effectiveness is often limited by substantial domain gaps and the practical challenges associated with acquiring target data, including cost and privacy constraints.
Consequently, DG has been developed to train models exclusively on source data while maintaining robust generalization to unseen target domains. This approach utilizes techniques ranging from data augmentation \cite{li2021simple,wang2022semantic,zeng2023foresee} to meta-learning \cite{li2018learning,shu2021open,chen2022compound,khoee2024domain} to learn domain-invariant features and prevent overfitting. Conventional DG (CDG), as depicted in Fig.~\ref{fig:sdg_task}(b), traditionally requires access to multiple source domains.

Recent feature-level augmentation methods, such as MixStyle \cite{zhou2024mixstyle}, and other feature interpolation techniques like FIXEDFE \cite{Lu2022FIXEDFE}, COCOA \cite{COCOA}, and XDomainMix \cite{XDomainMix}, have demonstrated substantial success in improving model generalization by synthesizing novel domain features. However, these approaches typically rely on interpolation across multiple real source domains. In this work, we address a different and more constrained scenario: we investigate whether feature interpolation remains effective when no additional real domains are available, compelling the model to learn by mixing features from a single source domain with synthetic data that are known to possess inherent artifacts and distributional biases.

\subsection{Single Domain Generalization} 

The task of SDG \cite{qiao2020learning,wang2021learning,yuan2023domain} has emerged as a particularly challenging yet practically significant subfield within DG. Unlike conventional multi-source DG, SDG is more realistic, as real-world applications often lack access to data from multiple domains. Its objective is to develop a generalized model trained on a single source domain that performs effectively across numerous unseen target domains. When applied to complex, dense-prediction tasks, this setting is often referred to as Single Domain Generalization in Object Detection (SDG-OD) \cite{wu2022sdgod} and Single Domain Generalization in Semantic Segmentation (SDG-SS) \cite{wang2022semantic}.
An intuitive approach to addressing SDG involves enhancing the diversity of single-source data through various data augmentation techniques. For instance, Carlucci et al. \cite{carlucci2019domain} employed a self-supervised learning strategy for data augmentation by segmenting images into tiles, shuffling them, and training the model to simultaneously classify the images and predict the correct order of the shuffled tiles. Similarly, Chen et al. \cite{chen2023center} proposed a center-aware adversarial augmentation technique that expands the source distribution by modifying source samples to distance them from class centers. Additionally, Lee et al. \cite{lee2024object} introduced an object-aware domain generalization approach, while Vidit et al. \cite{vidit2023clip} presented a novel semantic enhancement strategy utilizing a pre-trained vision-language model.

However, despite the success of data augmentation methods in image classification tasks, their adaptation to more complex computer vision tasks, such as object detection--which requires both classification and localization capabilities--remains challenging. Data augmentation in these contexts can inadvertently introduce noise, complicating model training \cite{liu2024unbiased}. Moreover, when only limited source domain data is available, even augmented pseudo-source domain data may lack the diversity necessary to effectively simulate real-world scenarios, because they do not incorporate knowledge beyond the original source domain.

\subsection{Synthetic Data-based SDG}
Recent advances in large-scale vision models underscore the potential of leveraging extensive prior knowledge embedded in foundational models pre-trained on internet-scale datasets. For instance, the C-Gap model \cite{vidit2023clip} attains SOTA performance in single-domain object detection by extracting pertinent semantic information from the CLIP model \cite{clip} using domain-specific text prompts. More recently, latent diffusion model \cite{ldm} has demonstrated a powerful ability to generate images with different styles by operating within latent spaces and incorporating text-guided techniques such as CLIP. By employing prompts and source domains to generate images across various domains, LDMs effectively simulate diversity in real-world scenarios \cite{danish2024improving,li2024prompt}. This approach mitigates the challenge of limited training data, offering a promising paradigm for generalization within a single domain.

However, our experiments reveal that while images generated directly using LDMs or existing data generation methods closely resemble real target domains visually, significant discrepancies persist in feature space. Utilizing such images without modification does not enhance, and may even impair, the model's generalization performance (refer to Section~\ref{sec:challenge}). Consequently, the primary motivation of this study is to explore how to effectively utilize synthetic data with inherent distribution bias for SDG tasks.

\begin{figure}[!t]
\begin{center}
    \centering
    \includegraphics[width=\linewidth]{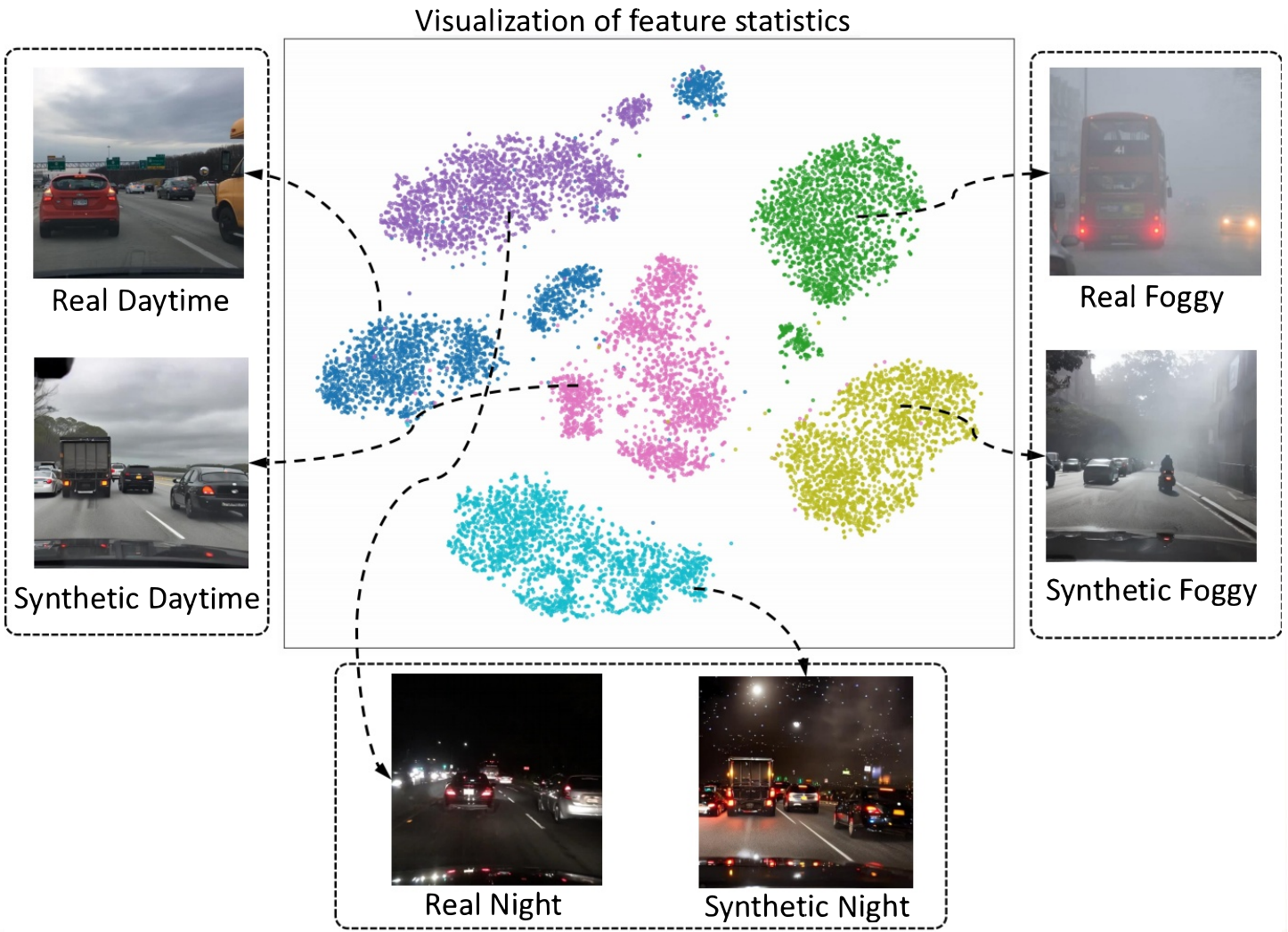}
    \caption{2D t-SNE visualization of image feature statistics for real and diffusion-generated domains.
}
    \label{fig:tsne_domain}
\end{center}
\end{figure}

\section{Motivation and Problem Formulation}

Diffusion models have emerged as powerful tools for enhancing visual representations, due to their high fidelity and ability to generate diverse cross-domain content. However, their application to SDG still presents considerable challenges. These models can efficiently synthesize multi-scene samples (e.g., foggy, nighttime, rainy, and snowy conditions) from a single source domain, offering new avenues to alleviate data scarcity. Yet existing research often overemphasizes the visual realism of synthetic data \cite{ldm,wang2024instancediffusion}, frequently at the expense of two critical properties: semantic consistency and distributional transferability between synthetic and real domains.

\begin{figure}[!t]
\begin{center}
    \centering
    \includegraphics[width=\linewidth]{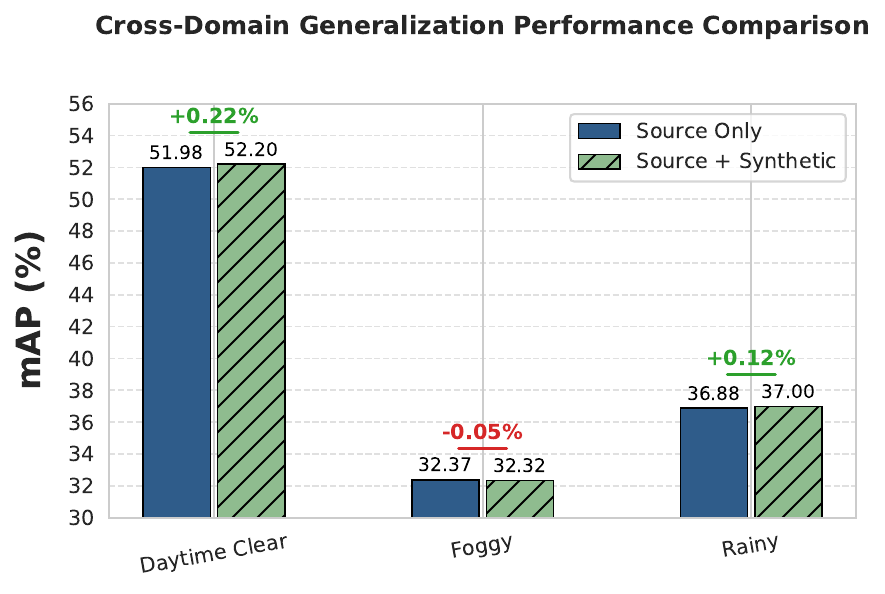}
    \caption{Impact of the synthetic domain on detector cross-Domain performance.
}
    \label{fig:comparison}
\end{center}
\end{figure}

\subsection{Core Challenges of Synthetic Data in Single Domain Generalization}\label{sec:challenge}
Data augmentation via diffusion models for SDG presents two primary challenges:

1) \textbf{Feature Space Discrepancy between Synthetic and Real Target Domains:}
As illustrated in Fig.~\ref{fig:tsne_domain}, despite the pixel-level similarity between diffusion-generated images and real images, these images exhibit excessive intra-domain clustering in feature space. Although visually diverse, the synthetic samples fail to adequately represent the distribution of the real target domain, hindering the model's ability to learn domain-invariant features.

2) \textbf{Artifacts and Distributional Shifts in Synthetic Data:}
The discrepancy in feature space can lead the model to learn spurious artifact patterns unique to the synthetic domain, such as non-physical fog density gradients, rather than generalizable semantic representations. Quantitative experiments using Faster R-CNN (Fig.~\ref{fig:comparison}) corroborate this issue: training with a direct fusion of the source domain (Daytime Clear) and diffusion-generated foggy/rainy data does not significantly improve performance on real target domains, and even degrades performance in some cases. This suggests that visual realism alone is insufficient to guarantee enhanced cross-domain generalization. Robust learning mechanisms are therefore needed to address biases introduced by synthetic data.

\begin{figure*}[!t]
\begin{center}
    \centering
    \includegraphics[width=\textwidth]{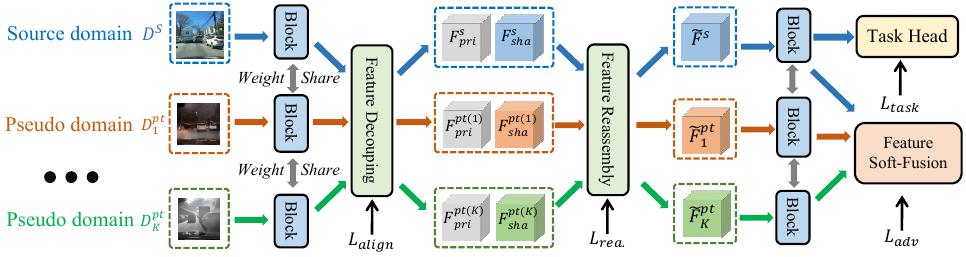}
    \caption{The overall DRSF framework. The backbone takes diverse pseudo-target domain data $\{\mathcal{D}^{pt}_i\}_{i=1}^K$, generated from a single source domain $\mathcal{D}^S$, as input. These features are then decoupled into primary features (domain-invariant features) and shared features (domain-specific features). Subsequently, a feature reassembly strategy is employed to mitigate interference arising from style variations. By integrating a multi-domain feature soft-fusion strategy, a continuous cross-domain feature space is constructed. The feature decoupling and reassembly are embedded within the backbone network blocks, while the feature soft fusion processes the output features from the backbone.
}
    \label{fig:overrall_framework}
\end{center}
\end{figure*}

\subsection{Formal Representation of SDG}

As depicted in Fig.~\ref{fig:sdg_task}(b), CDG requires models to learn domain-invariant representations from multiple source domains. This work addresses the more challenging SDG problem. Here, the model is trained solely on data from a single source domain, $\mathcal{D}^s$, and must generalize to $T$ unseen target domains, $\{\mathcal{D}_j^t\}_{j=1}^T$. Each target domain $\mathcal{D}^t$ follows an unknown distribution $\mathcal{P}_t(\mathbf{x}, \mathbf{y})$, where $\mathcal{P}_{\text {t}} \neq \mathcal{P}_{\text {s}}$. The objective of SDG is to learn a parameterized mapping $f_\theta: \mathcal{X} \rightarrow \mathcal{Y}$ that minimizes the risk function on the target domain:
\begin{equation}
\mathcal{R}_{\text {t}}\left(f_{\boldsymbol{\theta}}\right)=\mathbb{E}_{(\mathbf{x}, \mathbf{y}) \sim \mathcal{P}_{\text {t}}}\left[\mathcal{L}\left(f_{\boldsymbol{\theta}}(\mathbf{x}), \mathbf{y}\right)\right],
\end{equation}
where $\mathcal{L}(\cdot)$ denotes the task-specific loss function. Given that the target domain distribution $\mathcal{P}_{\text {t}}$ is inaccessible during training, the central challenge of SDG is to construct a sufficiently broad representation space from a single source distribution, enabling effective handling of potential domain shifts. Inspired by Ben-David et al.'s domain adaptation theory \cite{ben2010theory}, we recognize that the upper bound of the target domain error is related to both the source domain error and the domain discrepancy. Our key insight is that \textit{constructing a series of pseudo-target domains and establishing a continuous transition manifold in feature space can reduce the model's expected risk on unseen target domains}. As illustrated in Fig.~\ref{fig:sdg_task}(c), we specifically construct multiple pseudo-target domain distributions, $\{\mathcal{P}^{i}_{\text {pt}}\}_{i=1}^K$, to approximate potential target domains, and formulate a proxy optimization objective:
\begin{equation}
\small
\begin{split}
\min_{\boldsymbol{\theta}} \Big\{ & \frac{1}{K}\sum_{i=1}^{K} \mathbb{E}_{(\mathbf{x}, \mathbf{y}) \sim \mathcal{P}^{i}_{\text{pt}}} \left[ \mathcal{L}(f_{\boldsymbol{\theta}}(\mathbf{x}), \mathbf{y}) \right] \\
& + \lambda \cdot \frac{1}{K}\sum_{i=1}^{K}\mathcal{D}_{\phi}(\mathcal{P}_s, \mathcal{P}^{i}_{\text{pt}}) \Big\},
\end{split}
\end{equation}
where $\mathcal{D}_{\phi}(\cdot)$ symbolizes a parameterized generalization model to create a structured mapping in the source domain and several pseudo-target domains. Rather than merely considering mappings between individual domain pairs, this model establishes a continuous feature transformation space, effectively covering paths from the source domain to various potential target domains. Parameter $\lambda$ balances between DG and downstream task learning. Instead, we strive to build a feature space that is continuous and sufficiently broad and invariant to domains. Any unseen target domain should get approximately represented by the manifold of this feature space. Thus, covering distribution space allows the model to learn rich domain shift patterns and gain better generalization performance in unseen environments.

\section{Method}

\subsection{overview}\label{method_overview}

To overcome the challenges of SDG and fully exploit synthetic data, we introduce Domain Reassembly and Soft-Fusion (DRSF), a novel general training framework. As illustrated in Fig.~\ref{fig:overrall_framework}, DRSF adopts a four-stage collaborative learning paradigm-- generation, decoupling, reassembly, and fusion--to maximize the utility of synthetic data:

1) \textbf{Pseudo-target domain generation}: We leverage diffusion models to synthesize pseudo-target domain data with diverse stylistic variations. While preserving the semantic structure of the source domain $\mathcal{D}^S$, this process simulates domain shifts across different environmental conditions. Although this expands training data diversity, synthetic data still suffers from feature space bias and artifact patterns as analyzed in Section \ref{sec:challenge};

2) \textbf{Discriminative Feature Decoupling and Reassembly (DFDR)}: To mitigate limitations of synthetic data, we propose an embedded DFDR module within the backbone network. DFDR decomposes features into primary features (domain-invariant) and shared features (domain-specific).  By employing entropy-guided channel attention mechanisms, we selectively amplify synergistic channels while suppressing noisy ones. This dual process reduces artifact interference in synthetic data while strengthening cross-domain discriminative capabilities;

3) \textbf{Multi-Pseudo-domain Feature Soft-Fusion (MDSF)}: The MDSF module establishes smooth feature transitions between the source domain and multiple pseudo-target domains using soft label assignments and linear interpolation. By Combining with adversarial training, MDSF enables the model to learn domain-invariant representations that effectively cover the distribution space of potential target domains.

Unlike conventional approaches focusing solely on visual realism \cite{ldm,jia2024dginstyle}, DRSF addresses feature space biases through structural feature reassembly and continuous fusion. This turns synthetic data into useful tools that improve generalization. The framework's modular design allows seamless integration with existing UDA methods, making it broadly applicable beyond single-domain scenarios.

\subsection{Pseudo-target Domain Generation}\label{method_domain_gen}

In this section, we describe how LDMs are employed to generate pseudo-target domains that maintain semantic consistency while introducing stylistic diversity. This serves as the foundation for subsequent feature decoupling and fusion steps. The key advantages of using diffusion models for pseudo-target domain generation include: 1) Pixel-level fine-tuning is enabled through a combination of forward noise addition and gradual denoising processes; 2) Compared to GAN~\cite{goodfellow2020gan} and VAE~\cite{kingma2022vae}, diffusion models offer more stable style transformations without compromising semantic consistency; 3) They support a variety of conditional constraints to ensure high-quality generation. For our specific implementation, we utilize InstanceDiffusion~\cite{wang2024instancediffusion}, which leverages instance-level control to apply distinct style transformations to different objects within an image. This approach maximizes domain diversity while ensuring structural integrity is preserved.

As illustrated in Fig.~\ref{fig:sdg_task}(c), starting from a source domain dataset $\mathcal{D}^{s} = \{(\mathbf{x}_i^s, \mathbf{y}_i^s)\}_{i=1}^{N_s}$, we proceed to construct pseudo-target domain images:
\begin{equation}
  \mathbf{x}_i^{pt} = \mathcal{F}\bigl(\mathbf{x}_i^s, P^{pt}, C(\mathbf{x}_i^s)\bigr),
\label{eq:diff_gen}
\end{equation}

where $\mathcal{F}$ denotes the diffusion model, $P^{pt}$ represents the prompt for the target domain (e.g., ``nighttime foggy city street scene"), and $C(\cdot)$ imposes structural constraints (e.g., bounding box locations, edges). In the context of object detection, $\mathbf{y}_i^s = \{\mathbf{c}_j, \mathbf{b}_j\}_{j=1}^{N_b}$ comprises object categories $\mathbf{c}_j$ and bounding box coordinates $\mathbf{b}_j \in \mathbb{R}^4$.

\begin{figure*}[t]
\begin{center}
    \centering
    \includegraphics[width=\textwidth]{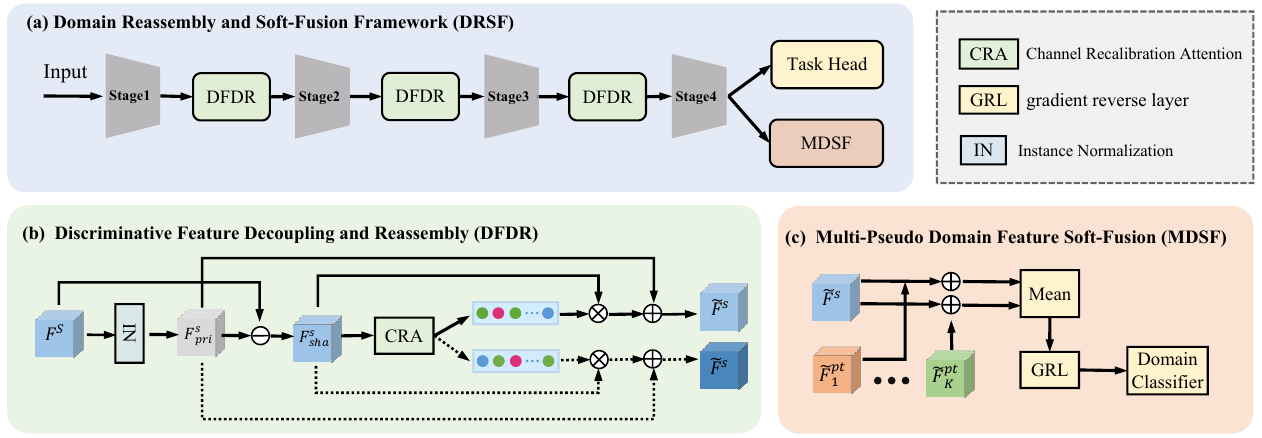}
    \caption{ (a) Illustration of the DRSF framework, where the DFDR module is embedded within the blocks of the backbone network, and the MDSF operates on the backbone's output. (b) Example diagram of the proposed DFDR. This module employs IN to decompose intermediate features, followed by CRA for feature reassembly. This feature reassembly strategy reduces interference arising from style variations. (c) The proposed MDSF. This module aims to achieve smooth fusion between the source and pseudo-target domains at the feature level through linear interpolation, thereby constructing a continuous cross-domain feature space.}
    \label{fig:overall_framework2}
\end{center}
\end{figure*}

To address the nuances of various visual tasks, we have tailored specific generative strategies: 

\begin{itemize}

\item \textbf{Semantic Segmentation:} To generate images with precise semantic layouts and diverse styles, we employ a style-prompting technique inspired by \cite{jia2024dginstyle}. We guide a ControlNet-based generator by first creating a base text prompt that concatenates the unique class names present in the source semantic mask (e.g., ``car, road, sky''). This prompt is then enriched by randomly fusing task-specific qualifiers, such as adverse weather conditions (e.g., ``foggy,'' ``snowy,'' ``rainy''), resulting in a composite prompt such as: ``A city street scene photo with car, road, sky... in foggy weather.'' This method ensures high fidelity to the original semantic structure while introducing varied stylistic appearances.

\item \textbf{Object Detection:} For object detection, we use a dual-prompting mechanism \cite{li2024godiff} to control both global scene characteristics and fine-grained object attributes.
Image-level prompts define the overall environment. We use Tag2Text \cite{huang2023tag2text} to extract descriptive tags from the source image (e.g., ``cityscape, street'') and merge them with target-domain descriptors (e.g., ``night, dark'') to construct a global prompt, such as ``a cityscape photo of a dark street at night.''
Object-level prompts provide detailed descriptions for individual instances. These are generated using a predefined template, e.g., ``A [\textit{object}] is [\textit{action}] on a [\textit{weather}] [\textit{scene}].'' Placeholders are filled with randomly selected attributes, yielding prompts such as ``A black car is parked on a foggy street.'' This hierarchical control produces a highly diverse and realistic pseudo-target dataset.

\end{itemize}

Following the strategy above, we construct $K$ distinct pseudo-target domain datasets $\{\mathcal{D}^{pt}_i\}_{i=1}^K$, each domain denotes as $\mathcal{D}^{pt} = \{(\mathbf{x}_i^{pt}, \mathbf{y}_i^{pt})\}_{i=1}^{N_{pt}}$. To address the key challenge of SDG when using synthetic data, the following sections will detail how we leverage the potential drawbacks of these pseudo-target domains to instead improve the model's generalization performance. 

\subsection{Discriminative Feature Decoupling and Reassembly}\label{method_reasembly}

This section introduces the Discriminative Feature Decoupling and Reassembly (DFDR) module. As illustrated in Fig.~\ref{fig:overall_framework2}(a), DFDR is embedded within the backbone network layers to process features.

To handle the features of the synthetic pseudo-domain data, we utilize Instance Normalization (IN) for feature decoupling. Originally introduced by Ulyanov et al. \cite{ulyanov2016instance} for style transfer, IN normalizes the features of each sample independently, effectively removing sample-specific style variations. Specifically, IN computes statistics for each instance across the spatial dimensions:
\begin{equation}
\mu_{nc}(F) = \frac{1}{HW}\sum_{h,w}F_{nchw},
\end{equation}
\begin{equation}
\quad \sigma_{nc}(F) = \sqrt{\frac{1}  {HW}\sum_{h,w}(F_{nchw}-\mu_{nc}(F))^2}.
\end{equation}

IN effectively disentangles content and style representations \cite{huang2017arbitrary}, enabling the distinction between structural information and stylistic noise in diffusion-generated images. This allows models to focus on learning domain-invariant features. Li et al. \cite{li2017demystifying} further showed that IN implicitly performs style normalization in neural feature spaces, separating content and style into distinct channel-based representations. 
Motivated by this theoretical basis and practical requirements, we developed an IN-based feature decoupling mechanism for synthetic data. This mechanism explicitly partitions features into domain-invariant primary features (capturing task-relevant structural information) and domain-specific shared features (encoding style variations), simultaneously addressing pseudo-domain feature biases and establishing a structural framework for subsequent reassembly.
Whereas prior work \cite{zhou2024mixstyle,li2017demystifying} often employ IN as a key step for style mixing or transfer, our focus is on the subsequent dynamics following this separation. We leverage IN as an explicit decoupling interface, enabling a subsequent, task-aware feature reassembly and cross-domain alignment process that is specifically designed to manage—rather than merely discard—style information from synthetic data to improve generalization.

As shown in Fig.~\ref{fig:overall_framework2}(b), the internal architecture of DFDR begins by decomposing the intermediate features $F \in \mathbb{R}^{N\times C\times H\times W}$ (with $N$ denoting batch size, $C$ channel count, $H$ spatial height, and $W$ spatial width) extracted from the backbone network into two distinct components:
\begin{equation}
\begin{aligned}
F_{pri} &= \gamma \odot \frac{F - \mu(F)}{\sigma(F)} + \beta, \\
F_{sha} &= F - F_{pri}.
\end{aligned}
\label{eq:dfdr_decouple}
\end{equation}

Specifically, $F_{pri}$ corresponds to domain-invariant primary features that encode task-relevant structural information, while $F_{sha}$ captures domain-specific shared features encompassing style and appearance variations. The affine parameters $\gamma$ and $\beta$ are learnable, and $\mu(F) \in \mathbb{R}^{N\times C}$ and $\sigma(F) \in \mathbb{R}^{N\times C}$ denote the spatially aggregated mean and standard deviation of the features, respectively.

We concurrently apply the decoupling process to features from the source domain and $K$ pseudo target domains, deriving a single set of domain-decoupled source features and $K$ corresponding sets for the pseudo target domains respectively:  
$\{F_{pri}^s, F_{sha}^s\}, \quad \{F_{pri}^{pt(i)}, F_{sha}^{pt(i)}\}_{i=1}^K$.

To enforce domain-invariant consistency in primary features, we employ the Maximum Mean Discrepancy (MMD) \cite{gretton2012mmd} loss to align the source domain's features with those of the pseudo target domains, ensuring cross-domain structural alignment.  
\begin{equation}
\mathcal{L}_{\mathrm{align}} = \frac{1}{K}\sum_{k=1}^K\text{MMD}(F_{pri}^s,\, F_{pri}^{pt(i)}).
\label{eq:mmd}
\end{equation}
By suppressing noise in synthetic imagery, this alignment mechanism fosters the development of robust domain-invariant feature representations within the latent space.

\subsubsection{Feature Reassembly}\label{Feature_Reassembly}  

While retaining primary features risks discarding valuable discriminative signals, shared features often encapsulate both informative and disruptive components. To balance noise suppression and discriminative channel utilization, we propose a \emph{style-recalibrated feature reassembly framework}. At its core lies the Channel Recalibration Attention (CRA), which performs fine-grained selection of salient features within shared channels. Unlike conventional attention schemes \cite{hu2018senet,lee2019srm}, our approach explicitly identifies ``gain" and ``interference" channels during training by analyzing the entropy of the network's feature input responses.  

The process begins with extracting style representations of shared features via global statistical modeling:  
\begin{equation}
Q_{n,c} = [\mu_{n,c}(F_{sha}) \,\parallel\, \sigma_{n,c}(F_{sha})],
\end{equation}
Here, $Q \in \mathbb{R}^{N \times C \times 2}$ encodes the global statistics $\mu_{n,c}$ (mean) and $\sigma_{n,c}$ (standard deviation) for the $c$-th channel of the $n$-th sample. Following this, we perform channel-wise linear transformations and batch normalization to refine the feature representations:  
\begin{equation}
T_{n,c} = W_c \cdot Q_{n,c},
\end{equation}
\begin{equation}
\hat{T}_{n,c} = \gamma_c\frac{T_{n,c} - \mu_c^{(T)}}{\sigma_c^{(T)}} + \beta_c,
\end{equation}
where $\mu_c^{(T)}$ and $\sigma_c^{(T)}$ represent the mean and standard deviation computed across the batch dimension, while $\gamma_c$ and $\beta_c$ denote learnable parameters. To obtain the channel-wise attention weights, we apply the Sigmoid function as follows:  
\begin{equation}
V_{n,c} = \sigma(\hat{T}_{n,c}) = \frac{1}{1 + e^{-\hat{T}_{n,c}}}.
\end{equation}
This yields an attention vector $V \in \mathbb{R}^{N \times C}$, where each element quantifies a channel's contribution to the discriminative task. Specifically, $V^s$ and $V^{pt}$ correspond to the channel recalibration attention vectors for the source domain and pseudo-target domain, respectively.

\paragraph{Feature Reassembly}  

To process the source domain's shared features $F_{sha}^s$, we employ the CRA attention vector $V^s$ to perform channel-wise fusion with primary features $F_{pri}^s$, generating two complementary residual components:  
\begin{equation}
\begin{aligned}
\widetilde{F}^s &= F_{pri}^s + V^s \odot F_{sha}^s,\\
\hat{F}^s &= F_{pri}^s + (1 - V^s) \odot F_{sha}^s.
\end{aligned}
\label{eq:fea_reassemble}
\end{equation}

The $\widetilde{F}^s$ emphasizes cooperative gain features by prioritizing high-attention channels, while $\hat{F}^s$ isolates interference features from low-attention channels. This dual decomposition enables distinct treatment of beneficial and detrimental channels during discriminative processing.  

\paragraph{Entropy Difference Supervision}  
To steer the optimization of feature reassembly, we propose a loss function based on predictive entropy differences:  
\begin{equation}
\begin{aligned}
\mathcal{L}_{\mathrm{rea}} = &\; \mathrm{ReLu}^{+}(H(\phi(\widetilde{F}^s)) - H(\phi(F_{pri}^s))) \\
&+ \mathrm{ReLu}^{+}(H(\phi(F^{s}_{pri})) - H(\phi(\hat{F}^{s}))),
\end{aligned}
\label{eq:loss_reassemble}
\end{equation}
where $H(\phi(F)) = -\sum_{k=1}^{N_K} p_{k}(F)\log p_{k}(F)$ quantifies the predictive entropy derived from feature $F$, with $\phi$ being the task-specific prediction head and $p_k(F)$ denoting the predicted probability for class $k$. The $\mathrm{ReLu}^{+}(x) = \ln(1 + e^x)$ operator smooths the regularization constraint.  
By comparing predictive entropies, this loss function promotes features that reduce uncertainty through synergistic interactions, while penalizing channels that introduce ambiguity. This entropy-driven supervision mechanism aligns with the feature space bias analysis in Section~\ref{sec:challenge}, thereby improving the utility of synthetic data during training.  

\subsection{Multi-Pseudo Domain-Based Feature Soft Fusion Training Framework}\label{method_domain_fusion}  

To establish a smoother cross-domain feature transformation space, this work introduces the Multi-Pseudo Domain Soft Fusion (MDSF) module (Fig.~\ref{fig:overall_framework2}(c)). By integrating feature-level interpolation and adversarial training, MDSF constructs a continuous feature space across multiple discrete pseudo-domains while learning domain-invariant discriminative representations.  

\subsubsection{Multi-Pseudo Domain Soft Fusion}  
In contrast to conventional input-space mixing that risks semantic distortion \cite{xu2020adversarial}, MDSF performs feature-level fusion on structured features from the DFDR module. This approach ensures precise integration as follows: First, we extract reassembled features from the $i$-th pseudo-target domain, then blend them with source-domain primary features to form a unified representation.  

\begin{equation}
    \widetilde{F}^{pt}_i = F_{pri}^{pt(i)} + V^{pt} \odot F_{sha}^{pt(i)},
\end{equation}
\begin{equation}
    F_i^{sf} = \lambda\,F_{pri}^s + (1-\lambda)\widetilde{F}^{pt}_i.
\label{eq:mixup_f}
\end{equation}

The mixing coefficient $\lambda$ is sampled from a symmetric beta distribution $\text{Beta}(\alpha_1, \alpha_2)$ with $\alpha_1 = \alpha_2$, where $\alpha=2.0$ ensures a balanced sampling distribution. This feature-level interpolation constructs a continuous feature path from the source domain to the $i$-th pseudo-target domain. To synchronize domain label consistency with feature blending, we apply corresponding label interpolation in parallel:  
\begin{equation}
    L_i^{sf} = \lambda\,L^s + (1-\lambda)\,L^{pt}_i,
\label{eq:mixup_label}
\end{equation}
where $L^s$ and $L^{pt}_i$ denote the domain labels for the source and pseudo-target domains, respectively. By employing soft labels, the adversarial training achieves smoother gradient distributions, thereby circumventing domain boundary conflicts that hard labels might induce.

Aiming to broaden domain coverage and mitigate domain-specific noise from individual pseudo domains, we average all interpolated features across source-pseudo domain pairs:  
\begin{equation}
\widetilde{F}^{sf} = \frac{1}{K}\sum_{i=1}^K F_i^{sf}.
\end{equation}

This approach effectively constructs a feature space centered at the source domain and radiating towards multiple pseudo-target domains. The corresponding domain labels are fused using the same strategy:  
\begin{equation}
\widetilde{L}^{sf} = \frac{1}{K}\sum_{i=1}^K L_k^{sf}.
\end{equation}

\subsubsection{Adversarial Training}  
To further encourage learning of domain-invariant discriminative features, we employ a GRL-based adversarial training framework \cite{ganin2015GRL, ganin2016domain}. A domain classifier $\mathcal{D}$ is designed to discriminate between source domain features, pseudo-target domain features, and their fused counterparts, while adversarial optimization eliminates domain-specific information through opposing objectives:  
\begin{equation}
\small
\begin{aligned}
    \mathcal{L}_{adv} = & \;\mathcal{L}_{\mathrm{CE}} \bigl(\mathcal{D}(\widetilde{F}^s),\, L^s\bigr) + \frac{1}{K}\sum_{i=1}^K \mathcal{L}_{\mathrm{CE}}\bigl(\mathcal{D}(\widetilde{F}^{pt}_i),\, L^{pt}_i\bigr) \\
    & + \mathcal{L}_{\mathrm{CE}}\bigl(\mathcal{D}(\widetilde{F}^{sf}),\, \widetilde{L}^{sf}\bigr),
\end{aligned}
\label{loss_adv}
\end{equation}
where $\mathcal{L}_{\mathrm{CE}}$ represents the cross-entropy loss. In this adversarial framework, the domain classifier $\mathcal{D}$ strives to distinguish features across domains, while the feature extractor learns domain-invariant representations via GRL. This dual process forces the model to prioritize cross-domain discriminative patterns while suppressing domain-specific noise. The MDSF module synergizes with the DFDR module: DFDR improves intra-domain feature quality through decoupling, whereas MDSF constructs a continuous cross-domain feature manifold. Together, they address the two critical challenges outlined in Sections \ref{sec:challenge}, thereby significantly boosting the model's domain generalization performance.

\subsection{Joint Optimization Objective}\label{method_objective}  
Building upon the DFDR and MDSF modules described earlier, we formulate a comprehensive optimization objective that balances task performance, feature quality, and domain generalization capabilities. Our total loss function is defined as:  
\begin{equation} \mathcal{L}_{\mathrm{total}} = \mathcal{L}_{\mathrm{task}}  
+\lambda_1 \mathcal{L}_{\mathrm{align}}  
+\lambda_2 \mathcal{L}_{\mathrm{rea.}}  
+\lambda_3 \mathcal{L}_{\mathrm{adv}}, \label{eq:overall_loss} \end{equation}  
where $\mathcal{L}_{\mathrm{task}}$ represents the downstream task-specific losses, such as detection classification/regression or segmentation pixel-wise cross-entropy, which ensure that the model has a discriminative foundation. $\mathcal{L}_{\mathrm{align}}$ (Eq. \ref{eq:mmd}) measures the domain feature distribution gap via MMD between the source and pseudo-target domains to foster cross-domain alignment. On the other hand, $\mathcal{L}_{\mathrm{adv}}$ (Eq. \ref{loss_adv}) comes from adversarial domain classification in MDSF to encourage domain-invariant feature learning for robust generalization. Furthermore, $\mathcal{L}_{\mathrm{rea.}}$ invokes DFDR's entropy difference regularization in \ref{eq:loss_reassemble} to reinforce salient feature channels while suppressing noise, thus enhancing feature discriminability. These will be controlled by the hyperparameters ${\lambda_1},{\lambda_2},{\lambda_3}$, which will trade off these complementary objectives. The complete end-to-end training procedure of our DRSF framework is summarized in Algorithm~1 of the Supplementary Material.

\subsection{Application and Extension}\label{method_extension}
During training, the model ingests both the source domain and multiple pseudo-target domains. The DFDR module refines intermediate features extracted by the backbone, while the MDSF module performs soft fusion. A joint optimization strategy then updates the backbone, DFDR module, task-specific head, and domain classifier until convergence is achieved. In contrast, the inference phase is streamlined, retaining only the backbone and DFDR module. Test images are processed to yield enhanced features $\widetilde{F}^t$, which are directly fed into the task-specific head for prediction. The MDSF module and interference pathways are deactivated during inference, minimizing computational overhead and ensuring efficient deployment.

\subsubsection{Task-specific Implementation}
The DFDR and MDSF modules are designed to flexibly accommodate diverse visual tasks, with their primary distinction lying in entropy metric computation strategies:

\textbf{\textit{Object Detection}}
For detection architectures, the DFDR module is embedded within the backbone network to assess entropy differences at the region level. Specifically, classification entropy is calculated for each Region of Interest (RoI) feature. The effectiveness of feature selection is evaluated using the average entropy across all foreground RoIs. In two-stage detectors such as Faster R-CNN \cite{ren2015fasterrcnn}, entropy computation occurs post RoI Pooling; for single-stage frameworks \cite{lin2017focal}, calculations are performed at foreground locations within feature maps.

\textit{\textbf{Semantic Segmentation}}
In segmentation tasks, pixel-level entropy differences are computed. Our approach calculates prediction entropy at every spatial location in the decoder's output, then averages these values across the entire feature map to derive a global entropy measure. This is formalized as:
\begin{equation}
\small
H(\phi(F)) = -\frac{1}{HW}\sum_{h=1}^{H}\sum_{w=1}^{W}\sum_{k=1}^{N_K}p_{k,h,w}(F)\log p_{k,h,w}(F),
\end{equation}
where $p_{k,h,w}(F)$ represents the predicted probability of the pixel at coordinate $(h,w)$ belonging to class $k$.

\textbf{\textit{Image Classification}}
For image classification, the architecture is simplified. The DFDR module is embedded within the backbone’s intermediate blocks, as in other tasks. The key difference lies in the entropy computation and fusion stages. The prediction entropy for $\mathcal{L}_{\mathrm{rea.}}$ is calculated at the image level based on the final softmax output of the classifier. The MDSF module performs soft fusion on the global feature vectors before they are fed into the fully connected classifier layer. This straightforward adaptation allows DRSF to be applied to standard classification networks.

\begin{table*}[t]
    \makeatletter\def\@captype{table}\makeatother
    \centering
    \caption{Performance comparison with SOTA SDG-OD methods under different weather conditions. \textsuperscript{\textdagger} denotes training on source and synthetic data.}
    \centering
    \begin{tabular}{l|c|c|cccc|c}
        \shline
        \multicolumn{1}{l|}{\multirow{1}{*}{Method}} & Venue & \makecell{Daytime \\ Clear}  & \makecell{Night \\ Clear} & \makecell{Dusk \\ Rainy} & \makecell{Night \\ Rainy} & \makecell{Daytime \\ Foggy} & \makecell{mPT$\uparrow$ (\%)} \\ \midrule
        F-RCNN \cite{ren2015fasterrcnn}  & NeurIPS~(2015) & 50.2 & 31.2 & 26.0 & 12.1 & 32.0   & 25.3  \\
        IBN-Net \cite{IBN-Net}   & ECCV (2018) & 49.7 & 32.1 & 26.1 & 14.3 & 29.6 & 25.5   \\
        IterNorm \cite{huang2019IterNorm}   & CVPR (2019) & 43.9 & 29.6 & 22.8 & 12.6 & 28.4  & 23.4  \\  
        ISW \cite{ISW}   & CVPR (2021) & 51.3 & 33.2 & 25.9 & 14.1 & 31.8 & 26.2   \\
        F-RCNN\textsuperscript{\textdagger} & NeurIPS~(2015)& 49.8 & 31.8 & 26.5 & 12.5 & 32.7 & 25.9 \\
        IBN-Net\textsuperscript{\textdagger} \cite{IBN-Net} & ECCV (2018) & 50.1 & 32.5 & 26.2 & 13.9 & 31.4 & 26.0 \\
        IterNorm\textsuperscript{\textdagger} \cite{huang2019IterNorm} & CVPR (2019) & 48.9 & 32.4 & 27.2 & 13.0 & 32.6 & 26.3  \\      
        ISW\textsuperscript{\textdagger} \cite{ISW} & CVPR (2021) & 51.0 & 33.5 & 26.1 & 14.3 & 31.3 & 26.3 \\ \midrule
        SW \cite{pan2019SW} & ICCV (2019) & 50.6 & 33.4 & 26.3 & 13.7 & 30.8 & 26.1    \\
        S-DGOD \cite{wu2022sdgod}   & CVPR (2022) & \underline{56.1} & 36.6 & 28.2 & 16.6 & 33.5 & 28.7  \\
        C-Gap \cite{vidit2023clip}   & CVPR (2023) & 51.3 & 36.9 & 32.3 & 18.7 & 38.5 & 31.6  \\
        SRCD \cite{rao2024srcd}  & TNNLS (2024) & - & 36.7 & 28.8 & 17.0 & 35.9 & 29.6  \\
        SHADE \cite{SHADE}& IJCV (2024) & - & 33.9 & 29.5 & 16.8 & 33.4 & 28.4 \\
        PDOL~\cite{li2024prompt}   & CVPR (2024) & 53.6 & \underline{38.5} & 33.7 & \underline{19.2} & \textbf{39.1} & \underline{32.6}  \\
        OA-DG \cite{lee2024object}   & AAAI (2024) & 55.8 & 38.0 & \underline{33.9} & 16.8 & 38.3 & 31.8  \\ 

        Ours & - & \textbf{57.7} & \textbf{40.2} & \textbf{39.8} & \textbf{20.2} & \underline{38.6} & \textbf{34.7} \\
        \shline
    \end{tabular}
    \label{table:dwd}
\end{table*}

\subsubsection{Integration with UDA Methods}
The proposed method functions as a plug-and-play component that seamlessly integrates with UDA techniques, establishing an integrated framework with generation augmentation, feature optimization, and domain adaptation capabilities. This is implemented through:
\begin{equation}
\min_{\theta} \Bigl(\mathcal{L}_{\mathrm{total}}(\mathcal{D}^{s}; \theta) + L_{\mathrm{uda}}\bigl(\mathcal{D}^{s}, \{\mathcal{D}_{i}^{pt}\}_{i=1}^K; \theta\bigr)\Bigr),
\label{eq:uda_loss}
\end{equation}
where $\mathcal{L}_{\mathrm{total}}$ represents the combined loss from Eq. (\ref{eq:overall_loss}), while $L_{\mathrm{uda}}$ encapsulates domain alignment or self-supervised objectives specific to UDA. The pseudo-target domain datasets $\{\mathcal{D}_{i}^{pt}\}_{i=1}^K$, functioning as either substitutes or complementary data to the target domain, substantially enhance the diversity coverage of target domain variations.

\section{Experiments}
This section validates the effectiveness of our DRSF framework through three fundamental vision tasks: object detection, semantic segmentation and image classification. We conduct comprehensive comparisons with SOTA SDG methods, perform component-wise ablation studies, analyze detection performance, and present visual interpretations.

\subsection{Single Domain Generalization for Object Detection (SDG-OD)}

\subsubsection{Experimental Setup}
\textit{\textbf{Datasets}} The experiments utilize the Diverse Weather Dataset (DWD)~\cite{wu2022sdgod}, a benchmark for evaluating SDG in object detection. This dataset encompasses five weather conditions for autonomous driving scenarios: Daytime-Clear, Night-Clear, Night-Rainy, Dusk-Rainy, and Daytime-Foggy. The Daytime-Clear subset contains 27,708 images (19,395 for training, 8,313 for testing). Night-Clear includes 26,158 images. The rain-based scenes (Dusk-Rainy and Night-Rainy) were synthetically generated from BDD-100k~\cite{yu2020bdd100k} images, yielding 3,501 and 2,494 images respectively. The foggy scenario includes 3,775 images. All datasets share seven common object categories: bus, bike, car, motor, person, rider, and truck. For fair UDA comparisons, we also incorporate Cityscapes \cite{cordts2016cityscapes} and FoggyCityscapes \cite{li2023AdvGRL} datasets.

\textit{\textbf{Implementation Details and Evaluation Metrics}}

In our experimental protocol, we consider only Daytime-Clear training data as the source domain, reserving other conditions for testing. To create pseudo-target domains, we leverage the InstanceDiffusion model \cite{wang2024instancediffusion} for its ability to apply style transformations while preserving instance-level structural integrity via bounding-box constraints. We generated 11{,}233 images across five virtual scenarios: Virtual Daytime-Clear, Night-Clear, Daytime-Foggy, Dusk-Rainy, and Night-Rainy. For reproducibility, example generation prompts, generated images, and code are provided in the \textbf{supplementary material}.
The MIC framework \cite{hoyer2023mic} will be our default UDA component.
All detection experiments are conducted in a Faster R-CNN \cite{ren2015fasterrcnn} architecture with a ResNet-101 backbone \cite{he2016resnet} and stochastic gradient descent optimally trained with a 0.01 initial learning rate (batch size 8, 36K iterations) and step decay at 24K/32K iterations. Training is performed on four RTX 3090 GPUs. According to standard evaluation protocols \cite{wu2022sdgod}, mean Average Precision mAP@0.5 is reported, and the mean Performance across Target domains (mPT) is computed for a comprehensive assessment.

\subsubsection{Comparison with State-of-the-Art SDG-OD Methods}

\begin{table*}[t]
\centering
\makeatletter\def\@captype{table}\makeatother
\caption{Per-class results (\%) for Daytime-Foggy and Dusk-Rainy conditions. \textsuperscript{\textdagger} denotes our reproduction.}
\centering
\resizebox{\linewidth}{!}{
\begin{tabular}{@{}l|cccccccc|ccccccccc@{}}
\shline
\multirow{2}{*}{\textbf{Method}} & \multicolumn{8}{c|}{\textbf{Daytime-Foggy}} & \multicolumn{8}{c@{}}{\textbf{Dusk-Rainy}} \\ \cmidrule{2-17}
& Bus & Bike & Car & Mot. & Pers. & Rider & Truck & mAP & Bus & Bike & Car & Mot. & Pers. & Rider & Truck & mAP  \\ \hline
F-RCNN~\cite{ren2015fasterrcnn}  & 30.7 & 26.7 & 49.7 & 26.2 & 30.9 & 35.5 & 23.2 & 31.9  & 36.8 & 15.8 & 50.1 & 12.8 & 18.9 & 12.4 & 39.5 & 26.6 \\
SW~\cite{pan2019SW} &30.6 &26.2 &44.6 &25.1 &30.7 &34.6 &23.6 &35.2 & 38.8 & 16.7 &50.1 &10.4 &20.1 &13.0 &38.8 & 26.3\\
IBN-Net~\cite{IBN-Net} & 29.9 & 26.1 & 44.5 & 24.4 & 26.2 & 33.5 & 22.4 & 29.6 & 37.0 & 14.8 & 50.3 & 11.4 & 17.3 & 13.3& 38.4 &26.1\\
IterNorm~\cite{huang2019IterNorm} & 29.7 & 21.8 & 42.4 & 24.4 & 26.0 & 33.3 & 21.6 & 28.5 & 32.9 & 14.1 & 38.9 & 11.0 & 15.5 & 11.6 & 35.7 & 22.8 \\
ISW~\cite{ISW} & 29.5 & 26.4 & 49.2 & 27.9 & 30.7 & 34.8 & 24.0 & 31.8 & 34.7 & 16.0 & 50.0 & 11.1 & 17.8 & 12.6 & 38.8 & 25.9\\
S-DGOD~\citep{wu2022sdgod}  & 32.9 & 28.0 & 48.8 & 29.8 & 32.5 & 38.2 & 24.1 & 33.5  & 37.1 & 19.6 & 50.9 & 13.4 & 19.7 & 16.3 & 40.7 & 28.2 \\
SRCD~\citep{rao2024srcd} & \underline{36.4} & 30.1 & 52.4 & 31.8 & 33.4 & 40.1 & \underline{27.7} & 35.9  & 39.5 & 21.4 & 50.6 & 11.9 & 20.1 & 17.6 & 40.5 & 28.8 \\
C-Gap~\citep{vidit2023clip}  & 36.1 & \underline{34.3} & 58.0 & 33.1 & 39.0 & \underline{43.9} & 25.1 & 38.5  & 37.8 & 22.8 & 60.7 & \underline{16.8} & 26.8 & 18.7 & 42.4 & 32.3  \\
PDOL~\citep{li2024prompt} & 36.1 & \textbf{34.5} & \underline{58.4} & \underline{33.3} & \underline{40.5} & \textbf{44.2} & 26.2 & \textbf{39.1}  & 39.4 & 25.2 & \underline{60.9} & \textbf{20.4} & 29.9 & 16.5 & 43.9 & 33.7  \\
DivAlign\textsuperscript{\textdagger} \cite{danish2024improving} & \textbf{36.9} & 30.8 & 56.9 & 32.2 & 32.9 & 38.1 & \textbf{29.1} & 36.7 & \underline{46.0} & \underline{27.3} & 60.0 & 13.7 & \underline{31.4} & \underline{24.2} & \textbf{48.8} & \underline{35.9}\\
\midrule
\textbf{Ours} & 35.4 &	32.9 &	\textbf{61.1}&	\textbf{34.2}&	\textbf{41.1}&	42.8&	22.7&	\underline{38.6}& \textbf{47.5} &\textbf{30.6}&	\textbf{70.6}&	14.6&	\textbf{39.4}&	\textbf{28.6}&	\underline{47.7}&	\textbf{39.8}\\
\shline
\end{tabular}
}
\label{table:weather_results_combined}
\end{table*}

\begin{table*}[t]
\centering
\makeatletter\def\@captype{table}\makeatother
\caption{Per-class results (\%) for Night-Clear and Night-Rainy. \textsuperscript{\textdagger} denotes our reproduction.}
\centering
\resizebox{\linewidth}{!}{
\begin{tabular}{@{}l|cccccccc|ccccccccc@{}}
\shline
\multirow{2}{*}{\textbf{Method}} & \multicolumn{8}{c|}{\textbf{Night-Clear}} & \multicolumn{8}{c@{}}{\textbf{Night-Rainy}} \\ \cmidrule{2-17}
& Bus & Bike & Car & Mot. & Pers. & Rider & Truck & mAP & Bus & Bike & Car & Mot. & Pers. & Rider & Truck & mAP \\ \hline
F-RCNN~\citep{ren2015fasterrcnn} & 37.7 & 30.6 & 49.5 & 15.4 & 31.5 & 28.6 & 40.8 & 33.5  & 22.6 & 11.5 & 27.7 & 0.4 & 10.0 & 10.5 & 19.0 & 14.5 \\
SW~\citep{pan2019SW} & 38.7 & 29.2 & 49.8 & 16.6 & 31.5 & 28.0 & 40.2 & 33.4 & 22.3 & 7.8 & 27.6 & 0.2 & 10.3 & 10.0 & 17.7 & 13.7\\
IBN-Net~\citep{IBN-Net} & 37.8 & 27.3 & 49.6 & 15.1 & 29.2 & 27.1 & 38.9 & 32.1 & 24.6 & 10.0 & 28.4 & 0.9 & 8.3 & 9.8 & 18.1 & 14.3\\
IterNorm~\citep{huang2019IterNorm} &  38.5 & 23.5 & 38.9 & 15.8 & 26.6 & 25.9 & 38.1 & 29.6 & 21.4 & 6.7 & 22.0 & 0.9 & 9.1 & 10.6 & 17.6 & 12.6 \\
ISW~\citep{ISW} &  38.5 & 28.5 & 49.6 & 15.4 & 31.9 & 27.5 & 41.3 & 33.2 & 22.5 & 11.4 & 26.9 & 0.4 & 9.9 & 9.8 & 17.5 & 14.1\\
S-DGOD~\citep{wu2022sdgod} & 40.6 & 35.1 & 50.7 & 19.7 & 34.7  & \textbf{32.1} & 43.4 & 36.6  & 24.4 & 11.6 & 29.5 & \underline{9.8} & 10.5 & 11.4  & 19.2  & 16.6 \\
SRCD~\citep{rao2024srcd} & \textbf{43.1} & 32.5 & 52.3 & 20.1 & 34.8 & 31.5  & 42.9 & 36.7 & 26.5 & 12.9 & 32.4 & 0.8 & 10.2 & \underline{12.5} & 24.0 & 17.0 \\
C-Gap~\citep{vidit2023clip} & 37.7 & 34.3 & 58.0 & 19.2 & 37.6 & 28.5 & 42.9 & 36.9 & \underline{28.6} & 12.1 & 36.1 & 9.2 & 12.3  & 9.6 & 22.9 & 18.7 \\
PDOL~\citep{li2024prompt} & 40.9 & 35.0 & 59.0 & \textbf{21.3} & \underline{40.4} & 29.9  & 42.9 & 38.5 & 25.6 & 12.1 & 35.8 & \textbf{10.1} & 14.2 & \textbf{12.9}  & 22.9 & 19.2 \\
DivAlign\textsuperscript{\textdagger} \cite{danish2024improving} & 41.4 & \underline{36.5} & \underline{60.4} & \underline{20.2} & 38.3 & \underline{32.0} & \textbf{45.9} & \underline{39.2} & \textbf{32.2} & \underline{13.0} & \textbf{44.0} & 2.5 & \underline{15.7} & 10.6 & \textbf{32.2} & \textbf{21.4}\\
\midrule
\textbf{Ours}  & \underline{41.9}&	\textbf{38.3}&	\textbf{65.2} &	16.4&	\textbf{43.9}&	30.7&	\underline{45.3}&	\textbf{40.2} & 28.4 &	\textbf{13.4}&	\underline{41.9} &	2.2&	\textbf{18.0}&	11.5&	\underline{25.7}&	\underline{20.2}\\
\shline
\end{tabular}
\label{table:weather_results_combined1}
}
\end{table*}

\textit{\textbf{Overall SDG-OD Results}}

\begin{figure*}[t]
\begin{center}
    \centering
    \includegraphics[width=\textwidth]{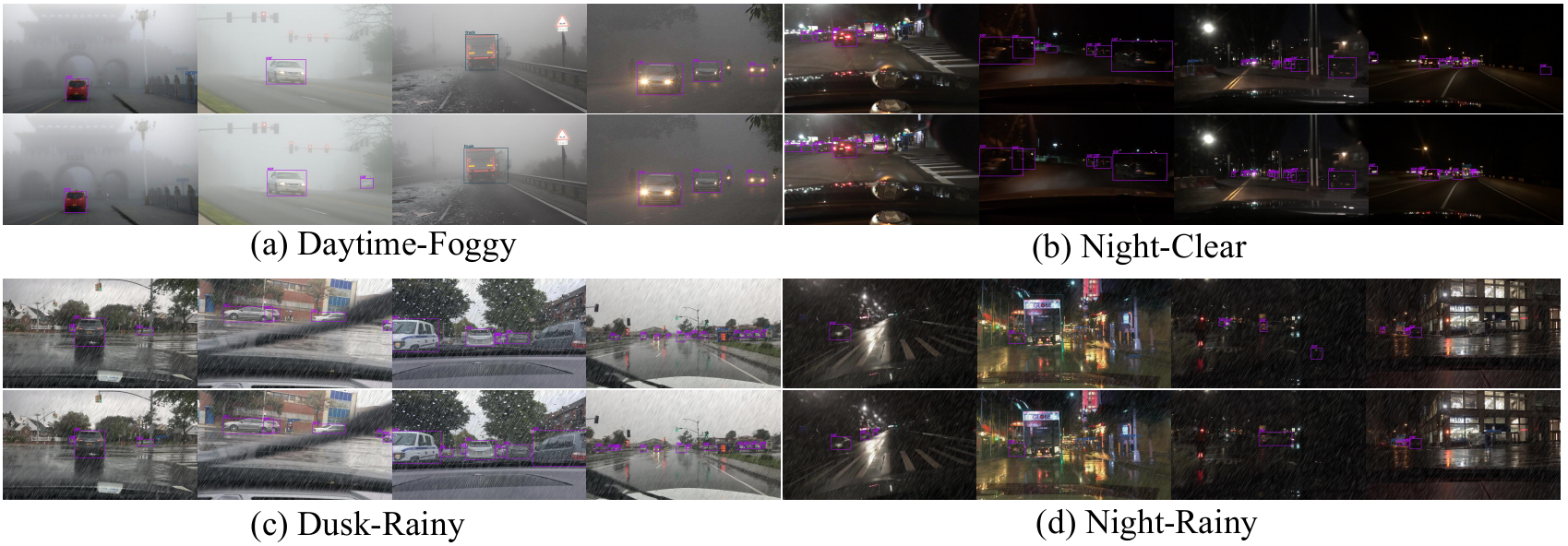}
    \caption{Qualitative comparison of object detection results between the baseline (vanilla Faster R-CNN, top row) and our method (bottom row) across diverse weather domains. Our method demonstrates improved accuracy compared to the baseline.}   \label{fig:predction_sdg}
\end{center}
\label{det_vis}
\end{figure*}

Table~\ref{table:dwd} reports a comprehensive comparison on DWD. We consider two baselines: source-only Faster R-CNN (25.3\% mPT) and a naive mixing baseline that jointly trains on the source plus our synthetic domains (F-RCNN\textsuperscript{\textdagger}, 25.9\% mPT). We further plug conventional DG algorithms into the same synthetic setup (\textsuperscript{\textdagger} denotes training on source+synthetic): IBN-Net\textsuperscript{\textdagger} reaches 26.0\% mPT, while IterNorm\textsuperscript{\textdagger} and ISW\textsuperscript{\textdagger} attain 26.3\% mPT. These modest gains confirm that simply pairing synthetic data with conventional DG yields limited benefits. 
In contrast, our DRSF achieves 34.7\% mPT, surpassing the best prior method (PDOL, 32.6\%) and OADG (31.8\%), while also improving source-domain performance to 57.7\%. This underscores the necessity of our feature-level decoupling-and-reassembly and multi–pseudo-domain soft-fusion design. 
Qualitative results in Fig.~\ref{fig:predction_sdg} further demonstrate more accurate localization and classification across diverse weather conditions.

\textit{\textbf{Daytime-Clear to Daytime-Foggy Scene}}
In addressing the domain shift from clear to foggy conditions, the left half of Table \ref{table:weather_results_combined} provides a category-wise performance analysis. Our method achieves an mAP of 38.6\%, significantly outperforming the SRCD (35.9\%). Notably, in the severely fog-affected ``Car" category, our approach attains 61.1\% AP, demonstrating the efficacy of our DFDR module in mitigating feature degradation caused by low-contrast fog. The strong performance (41.1\% AP) in ``Person" detection highlights our method's ability to preserve fine-grained discriminative features.

\textit{\textbf{Daytime-Clear to Dusk-Rainy Scene}}
For the challenging Dusk-Rainy scenario characterized by light attenuation and precipitation, results in Table \ref{table:weather_results_combined} (right) show our method achieves an mAP of 39.8\%, outperforming competing approaches. This validates our soft feature fusion strategy's capability to handle compound domain shifts. In critical categories like ``Car" (70.6\% mAP) and ``Person" (39.4\% mAP), we establish new performance benchmarks. For the rain-sensitive ``Bike" class, our 30.6\% mAP represents an 11.0\% improvement over S-DGOD, showcasing the synergy between DRSF mechanisms in complex environments.

\textbf{\textit{Daytime-Clear to Night-Clear Scene}}
The left half of Table \ref{table:weather_results_combined1} shows our method achieves an mAP of 40.2\%, surpassing PDOL~\cite{li2024prompt} (38.5\%) and S-DGOD (36.6\%). In nighttime scenarios, our method's generation of realistic headlight reflection features through diffusion models yields a record 65.2\% mAP in ``Car" detection. For ``Person" detection, our 43.9\% mAP exceeds PDOL~\cite{li2024prompt} by 3.5\%, addressing contour ambiguity under low-light conditions through enhanced feature robustness.

\textbf{\textit{Daytime-Clear to Night-Rainy Scene}}
In the most challenging Night-Rainy scenario (Table \ref{table:weather_results_combined1} right), our method maintains robustness with 20.2\% mAP. The 41.9\% car AP demonstrates effective noise suppression through multi pseudo domain feature fusion. While motorbike detection remains challenging (2.2\% AP), our approach shows 4.5× improvement over F-RCNN \cite{ren2015fasterrcnn}. Current limitations in rider detection (9.1\% AP) suggest future directions for disentangling precipitation artifacts from human textures. The difficulty in detecting motorbikes (Mot.) under night-rainy conditions is a common limitation across all tested methods, attributable to the compounded effects of low light, rain artifacts, and the objects’ small and complex structure. We provide a detailed failure-case analysis for this scenario in the Supplementary Material.

\subsection{Single Domain Generalization in Semantic Segmentation (SDG-SS)}

\subsubsection{Experimental Setup}
\textit{\textbf{Datasets}}
Following established SDG-SS protocols~\cite{wang2022semantic}, we utilize GTA~\cite{ros2016gta} as our source dataset, comprising 24,966 synthetic images. To evaluate cross-domain performance, three real-world autonomous driving datasets are employed. Cityscapes (CS)~\cite{cordts2016cityscapes} offers 500 German urban validation images. BDD100K (BDD)~\cite{yu2020bdd100k} provides 1,000 U.S. cityscape validation samples. Mapillary Vistas (MV)~\cite{neuhold2017mapillary} contains 2,000 globally sourced street images showcasing diverse environmental conditions. All experiments focus on 19 overlapping semantic categories common to these datasets.

\textit{\textbf{Implementation and Evaluation}}
Adhering to standard practices in SDG-SS, we evaluate our method using the DeepLabV2~\cite{chen2017deeplab} architecture with a ResNet-101 backbone pre-trained on ImageNet~\cite{deng2009imagenet}. For virtual domain generation, we adopt the Stable Diffusion 1.5-based pipeline introduced by Jia et~al.~\cite{jia2024dginstyle}. This pipeline leverages style-based text prompts in conjunction with the original ground-truth segmentation masks as topological priors, ensuring the preservation of precise semantic boundaries during style transfer. We created 6{,}000 virtual images with diverse weather conditions to serve as our pseudo-target domain. To simulate these conditions, we employed style prompts such as ``in the style of a foggy day,'' ``in the style of a rainy night,'' and ``in the style of a snowy winter day.'' Example generation prompts and generated images are provided in the \textbf{supplementary material}.

Based on the multi-resolution self-training strategy and training parameters suggested in MIC, we set the AdamW learning rates for the encoder and decoder at $6 \times 10^{-5}$ and $6 \times 10^{-4}$, respectively. Under the aligned SDG framework detailed in \cite{SHADE}, the model is trained for 40,000 iterations with a batch size of 2. Training and testing use 19 shared semantic categories, with the mean Intersection over Union(mIoU) over the 19 classes as the primary evaluation metric on the target datasets.

\subsubsection{Comparison with State-of-the-Art Methods}

Table~\ref{table:sdg-seg} compares DRDF against state-of-the-art SDG-SS methods, including classical techniques (IBN-Net \cite{IBN-Net}, ISW \cite{ISW}) and recent advances (SHADE~\cite{SHADE}, DGInStyle~\cite{jia2024dginstyle}). Using a ResNet-101 backbone trained on GTA, DRDF attains a 48.13\% average mIoU across three real-world target datasets, surpassing existing techniques by 1.5\%. Importantly, DRDF upholds strict SDG constraints, unlike methods requiring extra real-world data (marked with *) or unrealistic checkpoint selection (e.g., FSDR~\cite{huang2021fsdr}). Despite this, DRDF achieves top performance on all datasets: 50.19\% on Cityscapes, 43.50\% on BDD, and 50.72\% on MV. It outperforms SHADE~\cite{SHADE} and WEDGE~\cite{Wedge} by 2.86\% and 3.36\% in average mIoU, respectively. This validates our innovation of combining virtual domain reassembly and soft fusion strategies to learn domain-invariant representations, advancing SDG-SS.

\begin{table}[t]
    \makeatletter\def\@captype{table}\makeatother
    \centering
    \caption{Performance comparison with state-of-the-art methods on SDG-SS using ResNet-101. Models were trained on GTA5 and evaluated on Cityscapes, BDD100k, and Mapillary Vistas.}
    \centering
    \addtolength{\tabcolsep}{-3pt}
    \begin{tabular}{l|c|ccc|c}
        \shline
        \multicolumn{1}{l|}{\multirow{1}{*}{Method}} & Venue & \makecell{CS}  & \makecell{BDD} & MV & Avg. \\ \midrule
        
IBN-Net ~\cite{IBN-Net}& ECCV 18 & 37.37 & 34.21 & 36.81 & 36.13 \\

ISW ~\cite{ISW} & CVPR 21 & 37.20 & 33.36 & 35.57 & 35.38 \\

GTR* \cite{GTR} & TIP 21 & 43.70 & 39.60  & 39.10 & 40.80 \\
SAN-SAW \cite{SAN-SAW} & CVPR 22 & 45.33 & 41.18 & 40.77 & 42.43\\
FSDR \cite{huang2021fsdr}& CVPR 21 & 44.80 & 41.20 & 43.40 & 43.13 \\
AdvStyle \cite{AdvStyle} & NeurIPS 22 & 44.51 & 39.27 & 43.48 & 42.42 \\
WildNet* \cite{lee2022wildnet} & CVPR 22 & 45.79  &41.73 & 47.08 & 44.87 \\
WEDGE* \cite{Wedge} & ICRA 23 & 45.18  &41.06  &48.06 &44.77 \\
HRDA \cite{HRDA} & TPAMI 23 &39.63 & 38.69 &  42.21 & 39.18 \\
SHADE* \cite{SHADE} & IJCV 24 & 46.66 & 43.66 & 45.50 & 45.27\\
DGInStyle \cite{jia2024dginstyle} & ECCV 24 & 46.89 & 42.81 &50.19 & 46.63 \\ \midrule

Ours & - & \textbf{50.19} & \textbf{43.50} & \textbf{50.72} & \textbf{48.13} \\ \bottomrule

    \end{tabular}
    
    \label{table:sdg-seg}
    
\end{table}

\begin{figure}[t]
\begin{center}
    \centering
    \includegraphics[width=\linewidth]{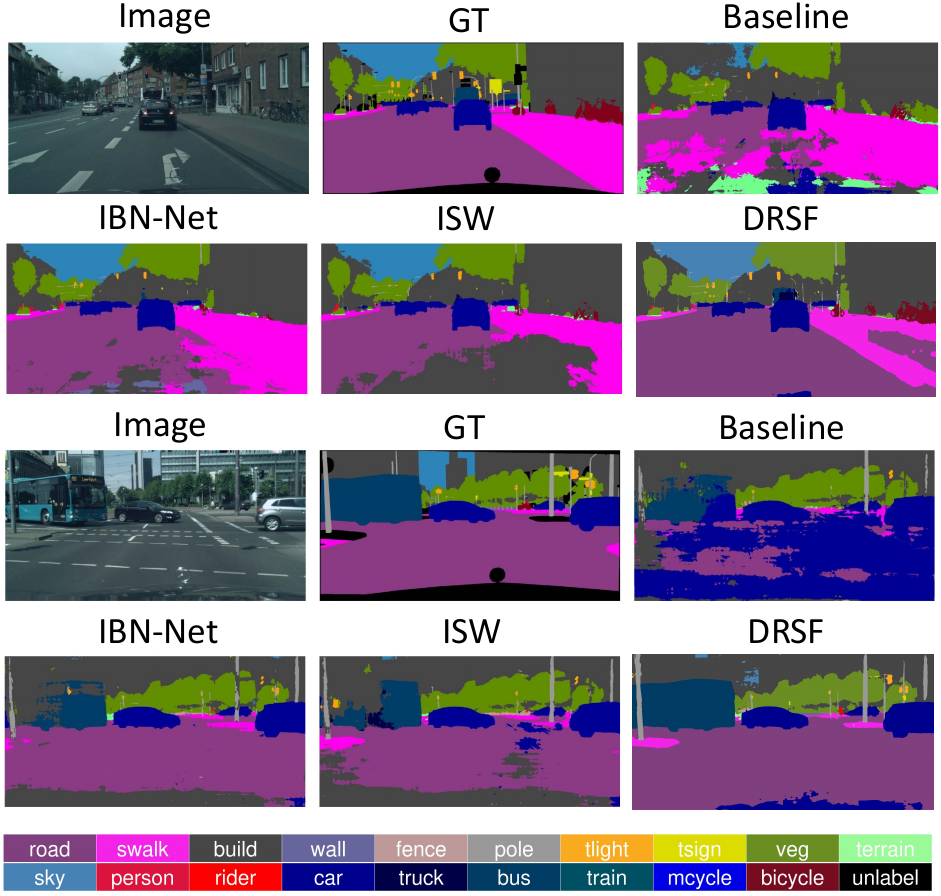}
    \caption{Qualitative comparison of semantic segmentation results between the baseline (vanilla DeepLabV2), IBN-Net, ISW, and our DRSF. Our method demonstrates improved accuracy compared to the baseline.}   \label{fig:predction_sdgss}
\end{center}
\end{figure}

Fig. \ref{fig:predction_sdgss} qualitatively demonstrates our method's effectiveness on the SDG-SS task. It compares DeepLabV2, IBN-Net, ISW, and our DRSF trained on GTA5 and tested on Cityscapes. DeepLabV2 suffers from severe domain shift, yielding fragmented predictions. While IBN-Net and ISW offer improvements, they struggle with large object structures (e.g., roads, sidewalks). In contrast, DRSF produces substantially more coherent segmentation masks with sharper boundaries and higher-fidelity representations of large structures, indicating superior learning of domain-invariant features.

However, the visualizations also highlight a common challenge for all methods, including ours. As seen in Fig. \ref{fig:predction_sdgss}, small and distant objects, such as riders and traffic signs, are often misclassified or missed entirely. This indicates that, while our method excels at capturing global semantic context and the structure of larger objects, improving robustness to fine-grained, small-scale elements remains an important direction for future research.

\begin{table}[t]
\small
\centering
\caption{Single domain generalization image classification results (\%) on PACS with backbone of ResNet-18. 
}
\label{pacs_single_result}
\addtolength{\tabcolsep}{-3pt}
\begin{tabular}{l|cccc|c}
\hline
Method  &  A & C & S & P & Avg  \\ \hline
RSC  \cite{huang2020rsc}& 73.40& 75.90 & 56.20 & 41.60 & 61.80 \\
L2D \cite{wang2021learning} & 77.26 & 78.27 & 53.40 & 49.06 & 64.50 \\
RSC+ASR \cite{fan2021asr}& 76.70 & 79.30 & 61.60 & 54.60 & 68.10 \\
Meta-Casual \cite{chen2023Meta-Casual} & 77.13 & 80.14 & 62.55&  59.60 & 69.86 \\
PDOL \cite{li2024prompt}   & 78.77 & \textbf{82.69} & \underline{62.94} & 60.09 & 71.12 \\ 
StyDeSty \cite{StyDeSty}& \underline{80.06} & 79.86 & 62.22 & \underline{63.25} & 71.35 \\ \midrule
Ours & \textbf{80.33} & \underline{82.67} & \textbf{64.92} & \textbf{63.52} & \textbf{72.86} \\        
\hline\end{tabular}

\end{table}

\begin{table*}[t]
  \centering
  \caption{Component-wise ablation studies on the SDG-OD benchmark. `mPT` denotes the mean Performance across the four Target domains.}
    \begin{tabular}{llllccccc|c}
    \toprule
    \multicolumn{2}{l}{DFDR} & \multicolumn{1}{l}{MDSF} & MIC   &  \multicolumn{1}{c}{Source} & \multicolumn{4}{c|}{Target} & \multicolumn{1}{c}{mPT $\uparrow$ (\%)} \\
    $\mathcal{L}_{\mathrm{align}}$ & $\mathcal{L}_{\mathrm{rea.}}$ & $\mathcal{L}_{\mathrm{adv}}$ &  & \makecell{D-Clear}& \makecell{N-Clear} & \makecell{D-Rainy} &\makecell{N-Rainy} &\makecell{D-Foggy} & \\ \midrule
        
      &    &   &   &  50.2 & 31.2 & 26.0 & 12.1 & 32.0 & 25.3 \\
        &  \checkmark  &   &  & 56.0 & 40.5 &32.8 &14.2 &35.1 & 30.7 \\
      \checkmark  &  \checkmark   &   &   & 57.0 & 40.5 &32.7 &14.9 &35.4 & 30.9 \\
            &    &  \checkmark  &  &  54.1 &35.8 &31.4 & 13.2 &36.2 & 29.2 \\
      \checkmark  &  \checkmark    &  \checkmark  &  &  \underline{57.5} & \textbf{40.5} & \underline{34.5} & \underline{16.2} & \underline{37.5} & \underline{32.2} \\

      &   &   &    \checkmark & 57.2 & 39.9 & 32.2 & 15.2 & 36.6 & 31.0 \\      
    \checkmark & \checkmark & \checkmark & \checkmark &  \textbf{57.6} & \underline{40.2} & \textbf{39.8} & \textbf{20.2} & \textbf{38.6} & \textbf{34.7} \\
    \bottomrule
    \end{tabular}%
  \label{tab:ablation_loss}%
\end{table*}%

\subsection{Single Domain Generalization in
Image Classification}
To validate the versatility of our DRSF framework beyond dense prediction, we extend our evaluation to the standard single domain generalization benchmark for image classification, PACS. This benchmark is well-suited for SDG, as it contains four distinct domains: Photo (P), Art Painting (A), Cartoon (C), and Sketch (S).

\subsubsection{Experimental Setup}
\textit{\textbf{Dataset and Protocol}} Following the standard leave-one-domain-out protocol, we train our model on a single source domain and evaluate its performance on the remaining three unseen domains. We report the mean accuracy across all four single-source training scenarios.

\textit{\textbf{Implementation Details}} For pseudo-target domain generation, we leverage a pre-trained Stable Diffusion 1.5 model. For each source image, we generate multiple stylized versions using prompts that combine the original class label with target style descriptors, such as ``a cartoon drawing of a [class\_name]'' or ``a sketch of a [class\_name]''. We create 5{,}014 virtual images in photo, art painting, cartoon, and sketch styles, aligned with the PACS benchmark. Example generation prompts and generated images are provided in the \textbf{supplementary material}. We use a ResNet-18 backbone pre-trained on ImageNet. The model is trained for 50 epochs using the Adam optimizer with a learning rate of $2 \times 10^{-2}$.

\subsubsection{Comparison with State-of-the-Art Methods}
As shown in Table~\ref{pacs_single_result}, our DRSF framework achieves a new SOTA mean accuracy of 72.86\% on the PACS benchmark. It outperforms prior methods, including those that leverage style augmentation or meta-learning. This result strongly supports that our core principles of discriminative reassembly and soft fusion are not task-specific but provide a general and effective mechanism for improving domain generalization. By effectively managing feature-level discrepancies in synthetic data, DRSF enhances the model's ability to learn domain-invariant representations, yielding robust performance even in the classic and highly competitive image classification setting.

\subsection{Ablation Study}
\subsubsection{Component Contributions}

Table~\ref{tab:ablation_loss} summarizes the performance impact of our core components, DFDR and MDSF. We use Faster R-CNN as the baseline, with the Daytime-Clear (D-Clear) domain as the source, and report the mAP on the source and four target domains, as well as the mPT.

The first row shows the baseline performance, achieving an mPT of 25.3\%. Introducing only the feature reassembly loss ($\mathcal{L}_{\mathrm{rea.}}$) from DFDR (second row) yields a significant 5.4\% absolute gain in mPT, primarily by improving performance in scenarios with varied lighting and weather conditions. This demonstrates that our entropy-driven mechanism is highly effective at selecting robust and discriminative feature channels.

When the full DFDR module is enabled by adding the alignment loss ($\mathcal{L}_{\mathrm{align}}$) (third row), the mPT shows a further modest gain to 30.9\%. While the average improvement appears small, the importance of $\mathcal{L}_{\mathrm{align}}$ is multi-faceted. First, it provides a stable gain on the most challenging domain (Night-Rainy, +0.7\%) and also improves source domain performance by 1.0\%, acting as a beneficial regularizer. Second, its primary role is to enable synergy with MDSF. As seen by comparing the fourth row (MDSF only, 29.2\% mPT) and the fifth row (DFDR + MDSF, 32.2\% mPT), the pre-alignment from DFDR unlocks a substantial +3.0\% mPT gain for the adversarial fusion process. This confirms that $\mathcal{L}_{\mathrm{align}}$ creates a better-structured feature space for subsequent modules.

To further validate the general effectiveness of our components, we conduct a similar ablation study on the SDG-SS task, with results shown in Table \ref{tab:ablation_loss_ss}. On this task, the alignment loss provides a more pronounced and consistent improvement, contributing 1.2\% to the average mIoU when added to the reassembly loss. This proves that all components, including the alignment term, are integral to our framework's success across different vision tasks.

Finally, integrating our full framework with an external UDA method like MIC further boosts the mPT to 34.7\%, confirming the compatibility and plug-and-play nature of our approach.

\begin{table}[t]
  \centering
  \caption{Component-wise ablation study on the SDG-SS benchmark.}
    \begin{tabular}{l|ccc|c}
    \toprule
    Method & CS & BDD & MV & Avg. \\ \midrule
    Baseline & 36.2 & 36.7 & 43.8 & 38.9 \\
    + $\mathcal{L}_{\mathrm{rea.}}$ & 45.4 & 41.0 & 47.1 & 44.5 \\
    + $\mathcal{L}_{\mathrm{rea.}} + \mathcal{L}_{\mathrm{align}}$ & 47.2 & 42.8 & 48.0 & 45.7 \\
    + $\mathcal{L}_{\mathrm{adv}}$ & 46.7 & 41.3 & 47.6 & 45.2 \\ \midrule
    Full Framework & \textbf{50.2} & \textbf{43.5} & \textbf{50.7} & \textbf{48.1} \\
    \bottomrule
    \end{tabular}%
  \label{tab:ablation_loss_ss}%
\end{table}

\begin{table}[htbp]
  \centering
  \caption{Ablation study of DFDR applied at different ResNet-101 backbone stages.}
    \begin{tabularx}{\linewidth}{l|>{\centering\arraybackslash}X>{\centering\arraybackslash}X>{\centering\arraybackslash}X>{\centering\arraybackslash}X>{\centering\arraybackslash}X}
    \toprule
    Method  & \makecell{DC}& \makecell{NC} & \makecell{DR} &\makecell{NR} &\makecell{DF}  \\ \midrule
        
      Baseline  &  50.2 & 31.2 & 26.0 & 12.1 & 32.0 \\
      + DFDR (res1)   & 56.0 & 39.5 &38.8 &\underline{18.7} &37.1 \\
      + DFDR (res2)   &  57.0 &39.8 &\underline{39.4} & 18.2 &36.2 \\
      + DFDR (res3)  &  \underline{57.1} & \textbf{40.5} & 37.5 & 17.2 & \underline{37.5}\\
      + DFDR (res123)  & \textbf{57.6} & \underline{40.2} & \textbf{39.8} & \textbf{20.2} & \textbf{38.6} \\ 
      + DFDR (res1234)  & 53.6 & 35.6 & 34.1 & 14.4 & 35.6 \\ 
    \bottomrule
    \end{tabularx}%
  \label{tab:ablation_layer}%
\end{table}%

\subsubsection{Network Layer Impact Analysis}

Table~\ref{tab:ablation_layer} details the impact of applying DFDR to different ResNet-101 residual blocks (res1-4), revealing three key findings: 1)\textit{ Layer-Specific Sensitivity}: Applying DFDR at a single layer yields significant gains, with different layers showing selective sensitivity to specific domain shifts. The res1 configuration (shallow features) notably improves the Night-Rainy (NR) scenario (+6.6\%), res2 excels in Dusk-Rainy (DR) (+13.4\%), while res3 boosts Night-Clear (NC) and Daytime-Foggy (DF) by 9.3\% and 5.5\%, respectively. This suggests that features captured at different levels exhibit varying sensitivities to different domain shifts. 2) \textit{Multi-Layer Complementarity}: Applying DFDR to the first three residual blocks (res123) achieves optimal performance, improving the source domain by 7.4\% and target domains by an average of 10.9\%. This multi-level configuration enables simultaneous feature decoupling and reorganization at varying abstraction levels: shallow layers handle textures and low-level features, mid-layers process object structure and contours, collectively building a hierarchical domain-invariant representation. 3) \textit{Deep Layer Applicability Limitations}: Extending DFDR to all layers (res1234) significantly degrades performance, echoing observations by Zhou et al. \cite{zhou2024mixstyle}. This is primarily because res4, located deep within the network, possesses highly semantic feature representations tightly coupled with category labels. Applying feature decoupling at this level disrupts the model's ability to capture high-level semantic information, reducing discrimination.

\begin{table}[!t]
  \centering
  \caption{Ablation study on the effect of UDA methods.}
    \begin{tabular}{llllccccc}
    \toprule
    Method  & \makecell{DC}& \makecell{NC} & \makecell{DR} &\makecell{NR} &\makecell{DF} \\ \midrule
        
      Baseline  &  50.2 & 31.2 & 26.0 & 12.1 & 32.0 \\
      + SADA   &  \underline{57.6} &39.3 &\textbf{40.8} & \textbf{20.7} &\underline{38.5} \\
      + HRDA   &  57.4 &\underline{39.6} & \underline{40.1} & 18.7 &37.5 \\
      + MIC  & \textbf{57.7} & \textbf{40.2} & 39.8 & \underline{20.2} & \textbf{38.6} \\ 
    \bottomrule
    \end{tabular}%
  \label{tab:ablation_UDA}%
\end{table}%

\subsubsection{Synergistic Effects with Different UDA Methods}

To validate the DRSF-UDA integration paradigm, Table~\ref{tab:ablation_UDA} reports results when combined with three representative UDA methods.

SADA \cite{chen2021sada}, addressing image- and instance-level shifts, performs best in rainy scenarios with our framework. This indicates the synthetic pseudo-target domain captures key rainy characteristics, enabling SADA’s scale-aware self-adaptation to learn robust target representations under complex conditions. HRDA \cite{hoyer2022hrda}, which fuses high-resolution details with low-resolution context via multi-resolution training, excels in Night-Clear. The pseudo-target domain preserves night-scene structure, allowing HRDA to extract richer contextual cues. MIC, which learns spatial context via masked image consistency, performs strongest in Daytime-Foggy. Its mask prediction mechanism leverages the pseudo-target domain to model deep spatial dependencies, mitigating fog-simulation limitations.

Performance differences across scenarios reflect variability in pseudo-target domain quality. Under lighting changes (e.g., Night-Clear), methods perform similarly; under multi-factor degradations (e.g., Night-Rainy), gaps widen, revealing both the limits of pseudo-target simulation and differing sensitivities of UDA methods to domain quality. Overall, these results validate our pseudo-target domain-driven adaptation paradigm: by generating high-quality pseudo-target domains and integrating advanced UDA, we achieve strong generalization without real target data.

\begin{table}[t]
  \centering
  \caption{Ablation study on the number and type of pseudo-target domains used for training. Performance is evaluated on the source domain (DC: Daytime-Clear) and four target domains (NC: Night-Clear, DR: Dusk-Rainy, NR: Night-Rainy, DF: Daytime-Foggy).}
    \begin{tabular}{l|ccccc}
    \toprule
    Method  & DC & NC & DR & NR & DF  \\ \midrule
    Baseline  &  50.2 & 31.2 & 26.0 & 12.1 & 32.0 \\
    + pseudo DC   & 55.6 & 39.6 & 36.9 & 18.3 & 37.3 \\
    + pseudo NC   &  55.0 & \textbf{40.2} & 37.0 & 18.5 & 36.5 \\
    + pseudo DR  &  55.3 & 40.0 & 36.5 & 16.2 & 35.5\\
    + pseudo NR  &  54.1 & 38.1 & 36.6 & \underline{19.8} & 35.4\\
    + pseudo DF  & 54.5 & 39.2 & 36.0 & 18.3 & \underline{38.1} \\ 
    \midrule
    + all  & \textbf{57.7} & \textbf{40.2} & \textbf{39.8} & \textbf{20.2} & \textbf{38.6} \\ 
    \bottomrule
    \end{tabular}%
  \label{tab:ablation_domain}%
\end{table}%

\subsubsection{Analysis on Pseudo-domain Generation}
\label{sec:generation_analysis}
To assess the impact of generated data on SDG performance, we analyze pseudo-domain generation across three factors: the number of synthetic styles, text-prompt phrasing, and generator seed randomness under the SDG-OD setting.

\textbf{Impact of the Number of Styles} We examine how the quantity of pseudo-target domains affects performance. Table~\ref{tab:ablation_domain} reports results for different combinations of five pseudo-domains, using Daytime-Clear as source. Key findings:
1) \textit{Multi-domain synergy}: Using all five ("+ all") yields the best results across test scenarios. On Dusk Rainy (DR), the full configuration (39.8\% mAP) surpasses the best single domain (pseudo-NC, 37.0\%), showing complementary coverage of environmental variations and a stronger representation space.
2) \textit{Cross-domain transfer}: Each single pseudo-domain improves not only its corresponding real target domain but also others. For example, pseudo-NC raises DR by 11 points over the baseline, indicating effective learning of domain-invariant features.
3) \textit{Source-domain enhancement}: Joint training on all pseudo-domains increases source performance from 50.2\% to 57.7\%, suggesting that the learned invariances refine source representations and avoid the typical source–target trade-off.

\begin{table}[t]
  \centering
  \caption{Sensitivity to prompt phrasing for the "foggy" style. Performance (mAP, \%) is evaluated on the source (DC) and two representative target domains (DF and DR).}
  \begin{tabular}{l|ccc}
    \toprule
    \textbf{Prompt} & \makecell{DC} & \makecell{DF} & \makecell{DR} \\
    \midrule
    "A foggy city street scene"     & 57.7 & 38.6 & 39.8 \\
    "A city street on a foggy day"  & 57.5 & 38.4 & 39.5 \\
    "Photo of a street in dense fog"& 57.6 & 38.5 & 39.7 \\
    \bottomrule
  \end{tabular}
  \label{tab:ablation_prompt}
\end{table}

\textbf{Sensitivity to Prompt Phrasing}
To ensure our method is not overly sensitive to the exact wording of text prompts, we conducted an experiment using three semantically similar but syntactically different prompts to generate the ``foggy" pseudo-domain. As shown in Table~\ref{tab:ablation_prompt}, the model's performance on both the source and target domains remains remarkably stable. The mAP scores exhibit minimal fluctuation, confirming that our framework is robust to minor variations in prompt phrasing, provided that the core stylistic concept is conveyed.

\begin{table}[h]
  \centering
  \caption{Impact of generator seed variability. We report mAP (\%) on the source and target domains across three full training runs with different seeds.}
  \addtolength{\tabcolsep}{8pt}
  \begin{tabular}{lccc}
    \toprule
    \textbf{Run} & \textbf{DC} & \textbf{DF} & \textbf{DR} \\
    \midrule
    Seed 1 & 57.7 & 38.6 & 39.8 \\
    Seed 2 & 57.8 & 38.5 & 39.9 \\
    Seed 3 & 57.6 & 38.7 & 39.6 \\
    \midrule
    \textbf{Mean} & \textbf{57.7} & \textbf{38.6} & \textbf{39.8} \\
    \textbf{Std.} & 0.10 & 0.12 & 0.15 \\
    \bottomrule
  \end{tabular}
  \label{tab:ablation_seed_v2}
\end{table}

\begin{table*}[ht]
\centering
\caption{Per-class Results from Daytime Clear to Night Clear. The results of DAOD methods are from \cite{wu2022sdgod}. }
\small
\label{tab:DAOD_rainy}
\begin{tabular}{l|c|ccccccc|c}
\toprule
Method & target &  Bus          & Bike          & Car           & Motor           & Person           & Rider     & Truck & \textbf{mAP} \\
\midrule
F-RCNN\cite{ren2015fasterrcnn}      &  \ding{56} & 37.7 & 30.6 & 49.5 & 15.4 & 31.5 & 28.6 & 40.8 & 33.5 \\
DAF \cite{DAF}     & \ding{52} & 36.2 & 29.1 & 49.3 & 16.0 & 33.1 & 29.3 & 40.2 & 33.3 \\
CT \cite{CT}     & \ding{52} & 34.1 & 22.1 & 46.4 & 12.8 & 26.5 & 19.8 & 31.5 & 27.6 \\
SW \cite{pan2019SW}     & \ding{52} & 34.2 & 23.6 & 48.0 & 13.4 & 26.4 & 23.7 & 37.5 & 29.5 \\
ICCR \cite{ICCR}   & \ding{52} & 36.1 & 23.2 & 48.9 & 15.5 & 29.1 & 23.8 & 39.4 & 30.9 \\
VDD \cite{VDD}    & \ding{52} & 35.4 & 29.6 & 49.8 & 14.5 & 31.3 & 28.0 & 39.9 & 32.6 \\
\midrule
Ours             &  \ding{56} & 41.9&	38.3&	65.2&	16.4&	43.9&	30.7&	45.3&	40.2
\\
Ours             &  \ding{52} & 38.6&	40.2&	63.2&	20.4&	44.6&	29.5&	40.3&	39.5\\
\bottomrule
\end{tabular}
\end{table*}

\begin{table*}[ht]
\centering
\caption{Comparison of UDA methods for Cityscapes to FoggyCityscapes adaptation (clear to foggy).}
\small
\label{tab:city2foggy}
\begin{tabular}{l|ccccccccc}
\toprule
\textbf{Methods} & Bus & Bicycle & Car & Motor & Person & Rider & Train & Truck & \textbf{mAP} \\
\midrule
SCDA \cite{zhu2019scdc} & 39.0 & 33.6 & 48.5 & 28.0 & 33.5 & 38.0 & 23.3 & 26.5 & 33.8 \\ 
DA-Faster \cite{DAF} & 49.8 & 39.0 & 53.0 & 28.9 & 35.7 & 45.2 & 45.4 & 30.9 & 41.0 \\ 
GPA \cite{xu2020GPA} & 45.7 & 38.7 & 54.1 & 32.4 & 32.9 & 46.7 & 41.1 & 24.7 & 39.5 \\ 
RPN-PR \cite{zhang2021rpn} & 43.6 & 36.8 & 50.5 & 29.7 & 33.3 & 45.6 & 42.0 & 30.4 & 39.0 \\ 
UaDAN \cite{guan2021UaDAN} & 49.4 & 38.9 & 53.6 & 32.3 & 36.5 & 46.1 & 42.7 & 28.9 & 41.1 \\  
AdvGRL \cite{li2023AdvGRL}& 51.2 & 39.1 & 54.3 & 31.6 & 36.5 & 46.7 & \textbf{48.7} & 30.3 & 42.3 \\ 
D-adapt \cite{jiang2021dadapt}& 42.8 & \textbf{42.4} & 56.8 & \textbf{35.2} &  42.8 &  48.4 &  37.4 & 31.5 & 42.2 \\
\textbf{Ours}  & \textbf{54.7} &	41.0&	\textbf{59.3}&	33.8&	\textbf{45.7} &	\textbf{48.9}&	27.8&	\textbf{34.8} & \textbf{43.5}

 \\ 
\bottomrule
\end{tabular}
\end{table*}

\textbf{Impact of Generator Seed Variability}
Finally, we evaluate the stability of our framework with respect to the inherent randomness in the diffusion model’s generation process. We execute the entire pipeline—from pseudo-domain generation to model training—three times with different global random seeds, and report the results in Table~\ref{tab:ablation_seed_v2}. The mean performance on both the source and target domains exhibits minimal standard deviation across runs. This robustness likely arises because diversity is primarily induced by applying a consistent style across thousands of varied source images, which outweighs the seed-induced pixel-level randomness.

\begin{figure*}[ht]
\begin{center}
    \centering
    \includegraphics[width=\textwidth]{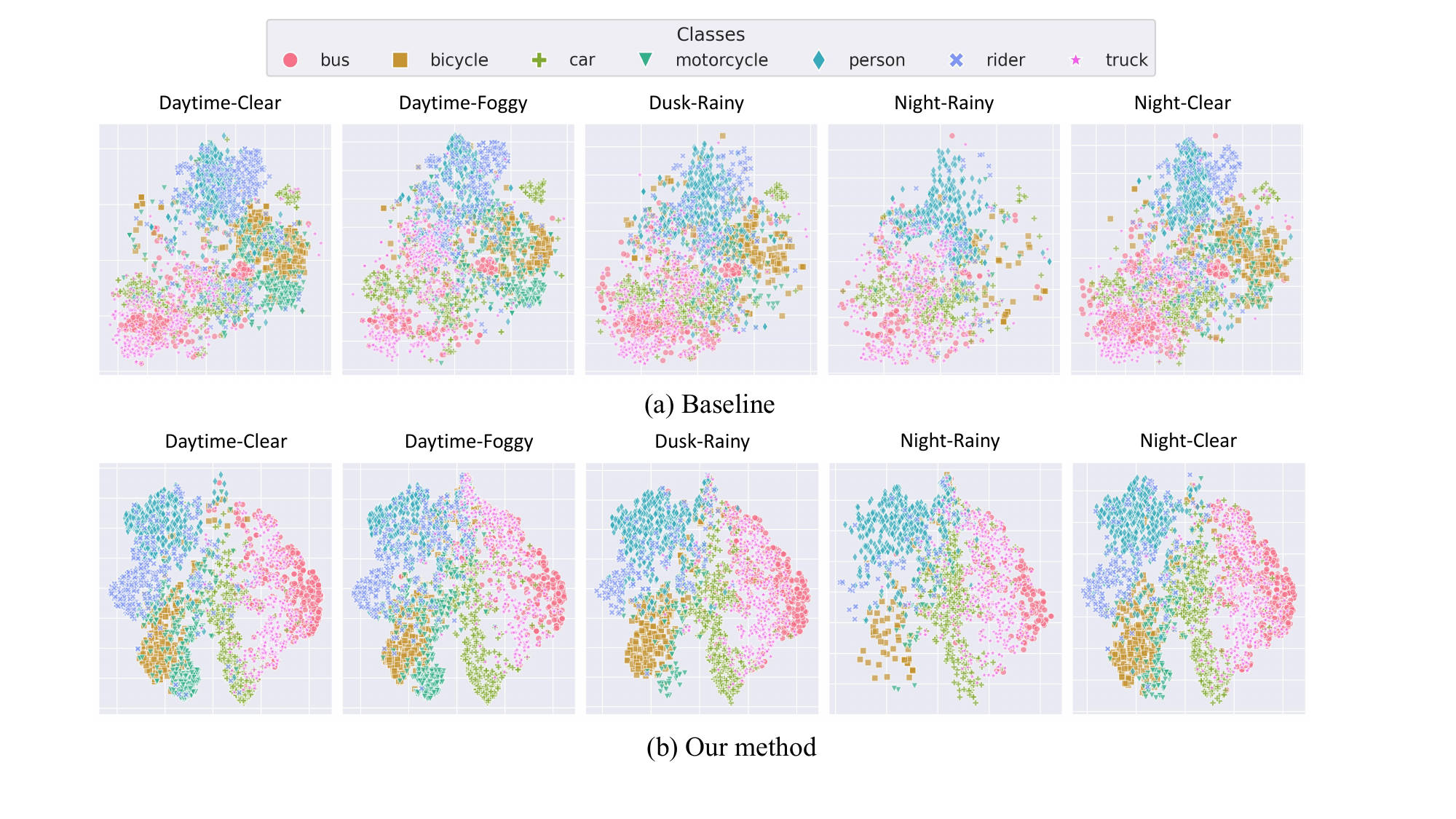}
    \caption{T-SNE visualization compares the feature distributions of the Baseline and our method for SDG-OD. Subfigures (a) and (b) depict the class feature distributions across five different domains for the Baseline and our method, respectively}
    \label{fig:tsne_sdg}
\end{center}
\end{figure*}

\subsection{Further Evaluation}

\subsubsection{Comparison with Domain Adaptation Methods}
To evaluate the proposed ``DRSF-UDA integration" framework, we conducted comparative experiments with UDA techniques across two standard cross-domain scenarios. Unlike traditional UDA approaches requiring access to target-domain samples, our investigation explores substituting or supplementing real target-domain data with synthetically generated pseudo-domain data.

\textbf{\textit{Daytime-to-Night Adaptation}}
As shown in Table \ref{tab:DAOD_rainy}, our method achieves a 40.2\% mAP without using any real nighttime target-domain data (denoted by ``\ding{56}"), outperforming all UDA baselines that rely on real target-domain data. The performance gains are particularly pronounced in ``Car" (15.4\% improvement) and ``Person" (12.6\% improvement) categories over the best UDA method. Counterintuitively, incorporating real target-domain data (marked ``\ding{52}") resulted in a slight performance drop to 39.5\% mAP. This unexpected trend suggests that noise and distribution biases in real target-domain samples may disrupt the domain-invariant feature learning from pseudo data. Notable degradation occurred in ``Bus" (-3.3\%) and ``Truck" (-5.0\%) categories, likely due to uneven nighttime lighting affecting large vehicle detection.

\textbf{\textit{Sunny-to-Foggy Adaptation}}
Table \ref{tab:city2foggy} demonstrates our method's 43.5\% mAP performance in foggy cityscape adaptation, surpassing competitive methods like AdvGRL and UaDAN. Five out of eight categories achieved state-of-the-art results, highlighting the DRSF framework's robustness against fog-induced visual degradation. Performance limitations are noted in the ``Train" category, primarily due to the sparse and morphologically varied train samples, which pose challenges for the feature decoupling strategy when dealing with large and rare objects. The underperformance in the ``Motor" category is likely caused by the ambiguity of contours induced by fog, which impedes the recovery of structural features. Despite these issues, the results confirm the dual effectiveness of DRSF: it functions both as an independent domain generalisation method and as an enhancement module for UDA approaches. This capability allows for superior performance with minimal dependence on real data from the target domain, highlighting the practical value of the proposed framework.

\begin{figure}[ht]
\begin{center}
    \centering
    \includegraphics[width=0.48\textwidth]{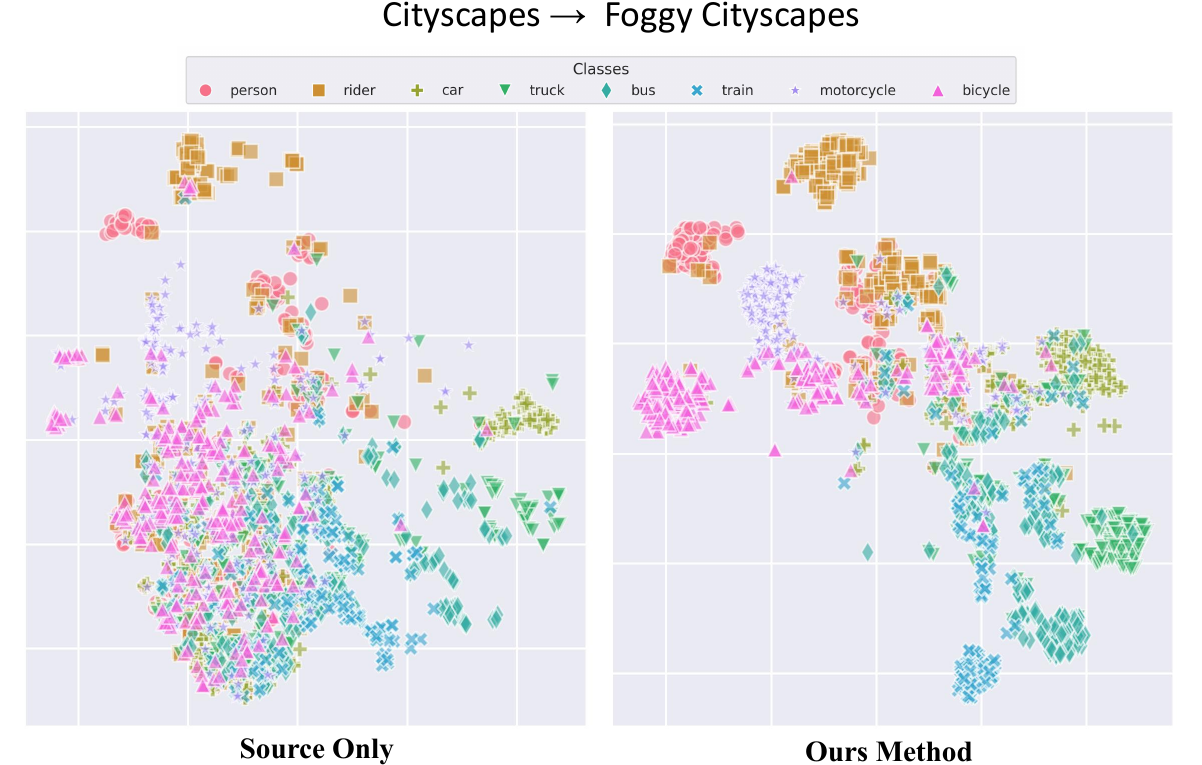}
    \caption{T-SNE visualization of feature embeddings in object detection domain adaptation, comparing source-only training with our proposed domain adaptation approach.}
    \label{fig:tsne_da}
\end{center}
\end{figure}

\subsubsection{Visual Analysis of Discriminative Model Features}

To intuitively demonstrate DRSF's improvements in feature representation, we visualized feature distributions across different domains using t-SNE, with detector-derived exemplar features representing category distributions. Fig.~\ref{fig:tsne_sdg} contrasts the feature spaces of baseline and DRSF models under the SDG-OD setting. The visualization highlights key differences: the baseline model's feature distribution exhibits distinct separation between source and target domains, with weak intra-class clustering. In contrast, DRSF achieves superior cross-domain alignment, maintaining the aggregation of same-category samples across domains. Especially in the sunny-to-foggy domain adaptation task (Fig.~\ref{fig:tsne_da}), the baseline model displays blurred inter-class boundaries and significant feature overlap, whereas DRSF effectively enhances feature discriminability, resulting in clear category boundaries. These visual results directly validate that DRSF successfully learns cross-domain shared discriminative representations while mitigating domain-specific interference.

\begin{figure}[t]
\begin{center}
    \centering
    \includegraphics[width=0.45\textwidth]{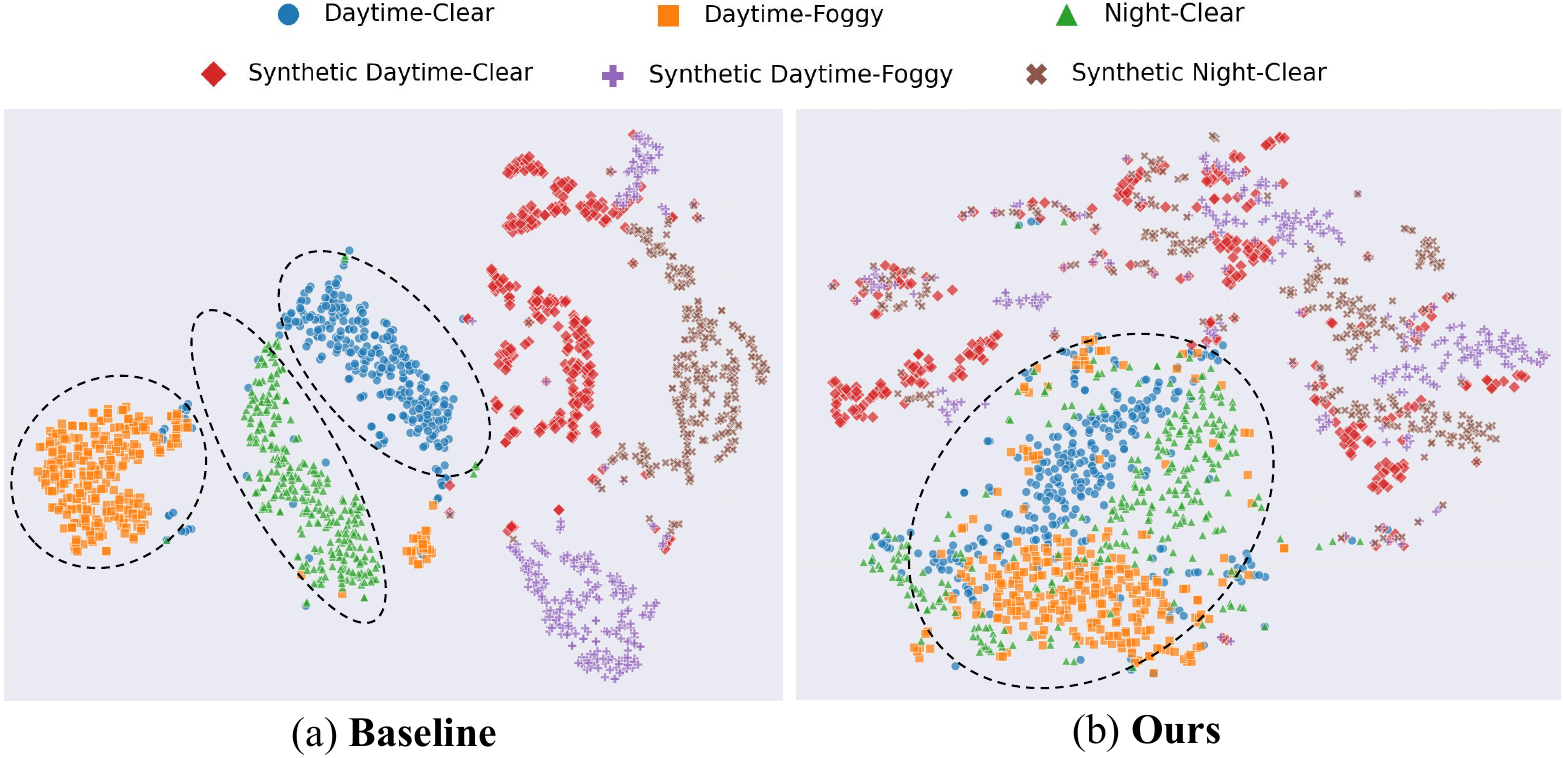}
    \caption{t-SNE visualization of domain features. (a) Baseline shows large inter-domain gaps. (b) Our method yields a unified cluster for real domains (Daytime-Clear, Daytime-Foggy, Night-Clear), demonstrating a domain-invariant representation under the SDG setting where no real target data are available during training. 
}
    \label{fig:tsne_ablation_domain}
\end{center}
\end{figure}

\subsubsection{Domain Feature Distribution Visualization}
\label{sec:vis_tsne}

This section presents an in-depth visualization analysis of feature representations learned by the DRSF framework. We use t-SNE to illustrate feature distributions extracted from the real source domain, unseen real target domains, and synthetic pseudo-target domains, as shown in Fig.~\ref{fig:tsne_ablation_domain}.

Fig.~\ref{fig:tsne_ablation_domain}(a) reveals that the baseline model (Faster R-CNN) forms distinct domain clusters in the feature space. The three real domains—Daytime-Clear (\textcolor[RGB]{36, 121, 181}{blue circles}), Daytime-Foggy (\textcolor[RGB]{254, 132, 25}{orange squares}), and Night-Clear (\textcolor[RGB]{59, 166, 60}{green triangles})—form mutually independent clusters, indicating a substantial domain shift. The synthetic domain data also exhibits a distinct distribution, suggesting that models lacking effective feature manipulation cannot leverage synthetic data to bridge domain gaps. Conversely, Fig.~\ref{fig:tsne_ablation_domain}(b) showcases the powerful effect of our DRSF framework. 
It is crucial to interpret this result from the perspective of SDG, where the model has \textit{no access to real target domains during training}. The primary objective is not to align synthetic data with an unknown real target, but to learn a representation that is robust to variations across all unseen real domains.
The key observation is that DRSF successfully aligns the features from all \textbf{real domains}. The features of the real source domain (\textcolor[RGB]{36, 121, 181}{blue circles}) and the two real target domains (\textcolor[RGB]{254, 132, 25}{orange squares} and \textcolor[RGB]{59, 166, 60}{green triangles}) are closely integrated, forming a single, unified cluster. This demonstrates that DRSF has learned a domain-invariant representation for real-world conditions, achieving the main goal of SDG.
Importantly, the synthetic pseudo-target domains move closer to the real-domain cluster yet remain partially distinct. This behavior is intentional and desirable in SDG: DRSF aligns content-related features while allowing style-related variations to differ, preventing the learned representation from being contaminated by non-photorealistic synthetic artifacts. Consequently, the synthetic data serves as a bridge that regularizes and expands the feature space to enhance robustness, without enforcing an unrealistic complete merger with real domains.

\begin{table}[ht]
    \footnotesize
    \centering
    \caption{ Performance and computational complexity on detection and segmentation.} 
    \addtolength{\tabcolsep}{-3pt}
     \begin{tabular}[h]{c ccc ccc}
     \toprule
     & \multicolumn{3}{c@{}}{Faster R-CNN}
     & \multicolumn{2}{c@{}}{DeepLabV2}\\
     \cmidrule(ll){2-4} \cmidrule(ll){5-6}
     & Baseline & OADG & DRSF & Baseline  & DRSF\\ 
     \midrule
     mAP($\uparrow$) & 50.2 & 55.8 & \textbf{57.4}  & - & -   \\
     mPT($\uparrow$) & 25.3 & 31.8 & \textbf{32.4}  & 38.9 & \textbf{48.1}  \\
     Params(M) & 51.96 & 52.29 & 52.54& 43.22&  43.92 \\
     FLOPs(G) & 778.4 & 779.8 & 779.4 & 1132.9 & 1151.3\\
     \bottomrule
     \end{tabular}
        
\label{tab:ablation_Complexity}
\end{table}

\subsubsection{Complexity Analysis}
We present a complexity study that includes both object detection (Faster R-CNN) and semantic segmentation (DeepLabV2), and we add comparisons with OADG~\cite{lee2024object} for detection. All models use a ResNet-101 backbone. Table~\ref{tab:ablation_Complexity} reports parameter counts and FLOPs measured under the same backbone and input settings.

For detection, DRSF improves mAP from 50.2 to 57.4 and mPT from 25.3 to 32.4 with only +1.1\% parameters and +0.1\% FLOPs. Compared to OADG, DRSF achieves superior accuracy at a similar computational cost. For segmentation, DRSF boosts cross-domain performance (mPT: 48.1 vs. 38.9) while introducing a small overhead over the DeepLabV2 baseline. These results corroborate that DRSF fits the prevailing “backbone + lightweight module” paradigm and delivers notable accuracy gains at negligible cost.

\begin{figure}[t]
\begin{center}
    \centering
    \includegraphics[width=0.5\textwidth]{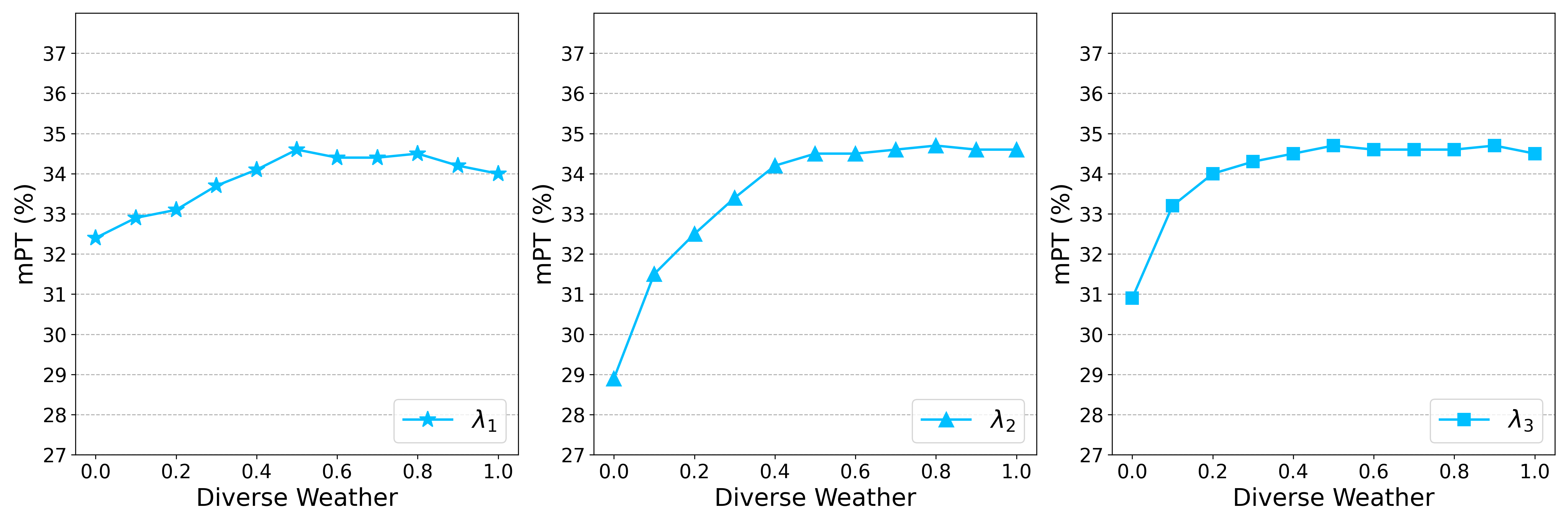}
    \caption{Sensitivity analysis to evaluate the impact of hyper-parameter $\lambda_1$, $\lambda_2$ and $\lambda_3$ on the performance of the SDG-OD task.}
    \label{fig:parameter_plot}
\end{center}
\end{figure}

\subsubsection{Hyperparameter Sensitivity Analysis}

We investigated the influence of three key hyperparameters controlling the feature alignment loss ($\lambda_1$), feature reassembly loss ($\lambda_2$), and domain adversarial loss ($\lambda_3$). Fig.~\ref{fig:parameter_plot} illustrates how the target domain mPT varies with these parameters. 

For $\lambda_1$, performance increases steadily as the weight grows from 0 to 0.5, but plateaus thereafter with marginal declines at higher values. This suggests moderate alignment effectively reduces domain structural differences without over-suppressing domain-specific feature learning. The $\lambda_2$ parameter shows the strongest impact, demonstrating nearly linear performance improvement across its range, underscoring the critical role of entropy-driven feature reassembly. The $\lambda_3$ parameter exhibits a ``rapid saturation" pattern: performance jumps sharply from 30.9\% to 34.7\% as $\lambda_3$ reaches 0.5, then stabilizes. This indicates sufficient adversarial strength enables the soft fusion mechanism to create a smooth feature transition space across pseudo-target domains. 

Based on this analysis, we recommend optimal settings of $\lambda_1=0.5$, $\lambda_2=0.8$, and $\lambda_3=0.5$, which achieve the best trade-off between performance enhancement and model stability. The results also highlight the importance of balancing adversarial learning intensity with domain adaptation objectives.

\section{Conclusion and Discussion}

This work addresses the critical challenge in SDG where synthetic data generated by LDMs often degrades model performance due to domain distribution discrepancies. We propose DRSF, a novel training framework that mitigates this issue through a feature-level domain reassembly and soft-fusion strategy.
Our extensive experiments demonstrate DRSF's effectiveness in improving cross-domain performance with minimal computational overhead. Moreover, DRSF's plug-and-play architecture enables seamless integration into existing pipelines, including extensions to unsupervised domain adaptation paradigms.

Limitations include current pseudo-domain generation's struggle with complex multi-degradation scenarios, which we propose to address through closed-loop optimization systems integrating model feedback. Future work should explore extending DRSF to multi-modal tasks such as instance segmentation~\cite{stformer}, and medical imaging~\cite{zhang2023medicalsg}. Additionally, combining DRSF with visual foundation models like~\cite{liu2024groundingdino} could leverage pre-trained model priors to further enhance generalization.

DRSF establishes a general approach to SDG by systematically addressing feature distribution bias and domain space coverage. Its plug-and-play design bridges the gap between synthetic data utilization and real-world domain adaptation needs, offering a practical solution for data-scarce environments while maintaining computational efficiency. This work not only advances SDG methodology but also provides a generalizable paradigm for DG research in open-world vision tasks.

\section*{Acknowledgements}
This work was supported by the National Natural Science Foundation of China under grants 62441618, 62276271, 62325604 and 62406100. Tianjin Natural Science Foundation under Grants No. 24JCQNJC00320.

\section*{Data Availibility}
The generated data using during the current study are available as follows: Object detection:
Semantic segmentation: 

The datasets analysed during the current study are available as follows: Diverse Weather \cite{wu2022sdgod}: \url{https://github.com/AmingWu/Single-DGOD} GTAV \cite{ros2016gta}: \url{https://download. visinf.tu-darmstadt.de/data/from_games/} CityScapes \cite{cordts2016cityscapes}: \url{https:// www.cityscapes-dataset.com/} BDD100K \cite{yu2020bdd100k}: \url{https:// bdd-data.berkeley.edu/} Mapillary \cite{neuhold2017mapillary}: \url{https://www. mapillary.com/dataset/vistas }. The generated data we used for SDG-OD and SDG-SS is available as follows: Synthetic Diverse Weather \url{https://drive.google.com/file/d/1F5Q7kO3uZCI6mWApW9e3-1yHVrBH37IN/view?usp=sharing} DGInStyle \cite{jia2024dginstyle} \url{https://drive.google.com/file/d/1e2wiwr5_wgCN3pLCQMOQvjRqQYtEozWy/view}

\section*{Code Availability}
The code is available at \url{https://github.com/fantasioly/DRSF}

\bibliography{ref(brief)}

\begin{appendices}

\end{appendices}

\end{document}